\documentclass{article}

\usepackage{PRIMEarxiv}
\usepackage[utf8]{inputenc} 
\usepackage[T1]{fontenc}    
\usepackage{hyperref}       
\usepackage{url}            
\usepackage{booktabs}       
\usepackage{amsfonts}       
\usepackage{nicefrac}       
\usepackage{microtype}      
\usepackage{amsmath}
\usepackage{amssymb}
\usepackage{graphicx}       
\usepackage{fancyhdr}       
\graphicspath{{media/}}     
\usepackage{booktabs}
\usepackage{multirow}
\usepackage{epstopdf}
\graphicspath{{./}}
\usepackage{grffile}

\title{\LARGE{FewTopNER: Integrating Few-Shot Learning with Topic Modeling and Named Entity Recognition in a Multilingual Framework}}

\author{
  Ibrahim Bouabdallaoui, Fatima Guerouate, Samya Bouhaddour, Chaimae Saadi, Mohammed Sbihi \\
  LASTIMI Laboratory, High School of Technology Salé, Mohammed V University in Rabat, Morocco \\
  \texttt{ibrahim\_bouabdallaoui@um5.ac.ma}
}

\begin{document}
\maketitle

\begin{abstract}
We introduce FewTopNER, a novel framework that integrates few-shot named entity recognition (NER) with topic-aware contextual modeling to address the challenges of cross-lingual and low-resource scenarios. FewTopNER leverages a shared multilingual encoder based on XLM-RoBERTa, augmented with language-specific calibration mechanisms, to generate robust contextual embeddings. The architecture comprises a prototype-based entity recognition branch—employing BiLSTM and Conditional Random Fields for sequence labeling—and a topic modeling branch that extracts document-level semantic features through hybrid probabilistic and neural methods. A cross-task bridge facilitates dynamic bidirectional attention and feature fusion between entity and topic representations, thereby enhancing entity disambiguation by incorporating global semantic context. Empirical evaluations on multilingual benchmarks across English, French, Spanish, German, and Italian demonstrate that FewTopNER significantly outperforms existing state-of-the-art few-shot NER models. In particular, the framework achieves improvements of 2.5–4.0 percentage points in F1 score and exhibits enhanced topic coherence, as measured by normalized pointwise mutual information. Ablation studies further confirm the critical contributions of the shared encoder and cross-task integration mechanisms to the overall performance. These results underscore the efficacy of incorporating topic-aware context into few-shot NER and highlight the potential of FewTopNER for robust cross-lingual applications in low-resource settings.
\end{abstract}


\section{Introduction}
Extracting structured insights from vast, unstructured textual data is a central challenge in natural language processing (NLP). Two core tasks—Named Entity Recognition (NER) and Topic Modeling (TM)—have traditionally been addressed in isolation, yet their integration promises a richer, more nuanced interpretation of text. NER targets the precise identification and categorization of entities (e.g., persons, organizations, and locations), which is critical for applications such as customer service automation, business intelligence, and real-time information systems in domains like healthcare and finance \cite{Pakhale2023Comprehensive}. In contrast, TM uncovers latent themes that characterize large document collections, supporting tasks such as content recommendation, summarization, and sentiment analysis.

Integrating NER and TM can yield significant benefits; recognizing entities within their broader thematic context enhances disambiguation and enriches downstream analyses such as trend detection and targeted marketing \cite{Yan2023Integrating}. However, the joint modeling of fine-grained entity details and macro-level topic structures is inherently challenging—especially in few-shot and resource-poor language scenarios where annotated data is limited \cite{Katz2023NERetrieve}. Traditional NLP models, even those based on recent pre-trained architectures like BERT and XLM-RoBERTa, often falter when required to generalize across languages and domains without extensive retraining \cite{Zaratiana2023GLiNER}.

To address these limitations, we propose FewTopNER, a novel architecture that unifies NER and TM within a multilingual, few-shot learning framework. At its core, FewTopNER employs the XLM-RoBERTa transformer as a shared encoder to generate robust, cross-lingual contextual embeddings. These embeddings serve as the foundation for two specialized branches: a prototype-based entity recognition module and a topic modeling module. The entity recognition branch leverages a Bidirectional LSTM and a Conditional Random Field to accurately delineate and classify entities even when training examples are scarce. In parallel, the topic modeling branch refines document-level thematic representations by combining language-specific probabilistic models with neural feature fusion.

A key innovation of FewTopNER is its Cross-Task Attention Module, which dynamically fuses information between the entity and topic branches. This bidirectional mechanism enriches entity representations with global semantic context and, conversely, enhances topic coherence by incorporating entity-level cues. To further bolster adaptability, FewTopNER integrates Model-Agnostic Meta-Learning (MAML), enabling rapid fine-tuning with minimal data and facilitating seamless extension to new languages or domains. Complementing this, an Active Learning Interface iteratively refines model predictions by targeting uncertain cases, ensuring continuous improvement in diverse application contexts.

In the sections that follow, we detail the FewTopNER framework and its underlying methodology, demonstrating how enriching entity recognition with topic modeling yields significant improvements in both tasks. Experimental evaluations across multiple languages reveal that FewTopNER not only achieves higher accuracy in entity extraction and topic coherence but also sets a new benchmark for NLP systems operating under data-sparse and linguistically diverse conditions \cite{Liang2024Large}.

\section{Related Work}

Named entity recognition (NER) has been a central task in natural language processing for decades, with early systems relying on rule-based and statistical approaches. Traditional NER systems \cite{Senapati2013Named-Entity, Danhui2014Rule-based, Sanglikar2011Named} exploited handcrafted rules, lexicons, and shallow syntactic features. While these systems achieved high precision on constrained domains, their recall was often limited due to the difficulty in generalizing beyond predefined dictionaries and patterns. To overcome these limitations, feature-based supervised methods, including Support Vector Machines (SVM) \cite{Lin2006Chinese, Paper2015A} and Conditional Random Fields (CRF) \cite{Guodong2011Geospatial, Tan2023Named}, emerged as the next generation of solutions, reducing the need for manual rule engineering but still relying heavily on carefully designed features.

\paragraph{Deep Neural Approaches.}  
Recent advancements in deep learning have transformed Named Entity Recognition (NER), shifting from feature-engineered models to end-to-end neural architectures. Recurrent Neural Networks (RNNs) have been widely used for sequence labeling tasks, demonstrating superior performance in capturing contextual dependencies within text \cite{Wu2017Clinical}. Meanwhile, Convolutional Neural Networks (CNNs) have been explored as an alternative approach, leveraging local context and character-level features to enhance entity recognition, as seen in biomedical NER applications \cite{Zhu2017GRAM}. More recently, transformer-based models such as BERT and XLNet have set new benchmarks in NER by leveraging large-scale pretraining to generate contextualized word embeddings, outperforming traditional RNN- and CNN-based models \cite{Lothritz2020Evaluating}. However, these models require substantial amounts of annotated data to avoid overfitting and achieve optimal performance, making them less effective in low-resource settings \cite{Ahmed2024Enriching}.

\paragraph{Transfer Learning and Domain Adaptation in NER.}  
To mitigate the data dependency of deep neural models, transfer learning has emerged as a key strategy for improving NER performance in low-resource domains. This approach enables knowledge transfer from resource-rich domains to enhance entity recognition in data-scarce scenarios \cite{Shah2021Multi-Task}. Techniques such as parameter-sharing and fine-tuning have been employed to optimize representations across different domains, improving adaptability while minimizing the need for extensive labeled data \cite{Widyawan2024A}. However, conventional transfer learning methods often assume that source and target domains share the same label space (homogeneous transfer) or require a sufficient number of annotated examples to bridge the distributional gap in heterogeneous settings. Recent advancements in adversarial domain adaptation have attempted to address these challenges by learning domain-invariant representations, effectively enhancing generalization across diverse datasets \cite{Zhu2024Chinese}. Despite these improvements, challenges persist in adapting NER models to domains with extremely scarce target data, highlighting the need for more efficient domain adaptation techniques.

\paragraph{Few-Shot Learning for NER.}  
Few-shot learning has become a crucial area of research for Named Entity Recognition (NER), enabling models to generalize to new entity types with minimal annotated data. Unlike traditional sequence labeling methods, few-shot NER presents unique challenges due to the variability of entity occurrences within a sentence and the absence of a predefined entity class set \cite{DeLichy2021Meta-Learning}.
To address these challenges, meta-learning techniques such as FewNER have been proposed. FewNER introduces a task-adaptive training approach that partitions the network into task-independent and task-specific components, allowing for rapid adaptation with minimal data while mitigating overfitting \cite{Li2023Few-Shot}. Additionally, decomposed meta-learning frameworks have been introduced to sequentially optimize span detection and entity classification, improving generalization across domains \cite{2022Decomposed}. Other advances include the integration of knowledge graphs to enhance prototype-based few-shot NER models \cite{Zhang2024KCL} and contrastive learning to improve entity cluster separation in low-resource settings \cite{Huang2022COPNER}. Despite these advancements, few-shot NER continues to face challenges in handling complex entity dependencies and adapting to highly diverse domains, necessitating further innovations in adaptive learning and self-supervised techniques.

\paragraph{Meta-Learning Approaches in NER.}  
Meta-learning has gained significant traction in natural language processing (NLP), particularly for Named Entity Recognition (NER), where labeled data is often scarce. One of the most notable applications of meta-learning in NER is MetaNER, which combines meta-learning with adversarial training to develop a robust and generalizable sequence encoder \cite{li2020metaner}. MetaNER is trained across multiple source domains, explicitly simulating domain shifts during training, and can adapt rapidly to new domains with minimal labeled data. By leveraging adversarial training, it enhances model generalization while mitigating overfitting issues. However, MetaNER requires updating the entire network during adaptation, which can be computationally intensive.
Beyond MetaNER, other meta-learning approaches for NER have emerged, including FewNER, which decomposes the meta-learning process into task-independent and task-specific components, reducing the risk of overfitting and improving adaptation efficiency \cite{Li2023Few-Shot}. Additionally, adversarial learning techniques have been explored to further enhance robustness against domain shifts and noisy data, demonstrating improvements in model generalization \cite{Fu2021Exploiting}. These advancements highlight the growing potential of meta-learning in NER, though challenges remain in making these methods more computationally efficient and scalable for real-world applications.

\paragraph{Prototypical Networks and Model Fusion.}  
Prototypical networks have become a fundamental approach in few-shot Named Entity Recognition (NER), leveraging token-level representations to construct class prototypes and using a distance metric—typically cosine similarity—for classification. ProtoNER exemplifies this paradigm, demonstrating strong adaptability in incremental learning scenarios where new entity classes can be incorporated with minimal additional data \cite{Fritzler2018Few-shot}. To improve upon standard prototypical networks, researchers have introduced SpanProto, a two-stage span-based prototypical network that refines entity boundary detection and enhances classification accuracy \cite{Wang2022SpanProto}. Additionally, EP-Net addresses prototype dispersion by aligning entity spans in a projected embedding space, leading to improved few-shot NER performance \cite{Ji2022Few-shot}.
Beyond standalone prototypical networks, researchers have explored model fusion techniques to mitigate overfitting in few-shot settings. Recent work has introduced logit fusion and differentiation fusion, which combine multiple model outputs to correct boundary detection and entity classification errors \cite{gong2021few}. These fusion strategies help stabilize model predictions by integrating diverse representations, thereby improving overall robustness. Furthermore, HEProto, a hierarchical enhancing prototypical network, employs multi-task learning to jointly optimize span detection and type classification, ensuring better entity type differentiation \cite{Chen2023HEProto}. Such advancements underscore the growing importance of hybrid approaches in refining few-shot NER models.

\paragraph{Our Positioning.}  
While existing approaches such as ProtoNER and MetaNER have made significant strides in addressing the challenges of few-shot NER and domain adaptation, they typically treat entity recognition and topic modeling as separate tasks. In contrast, our proposed FewTopNER integrates robust few-shot entity recognition with semantically rich topic modeling through cross-task attention and language-specific calibration. This integrated approach not only improves entity recognition accuracy by leveraging topic context for disambiguation but also produces more coherent topic representations, as evidenced by improved normalized pointwise mutual information (NPMI) scores. Furthermore, by retaining task-independent representations and updating only a small set of task-specific parameters, FewTopNER mitigates overfitting and enhances computational efficiency during adaptation.
Prior work in NER has evolved from rule-based and statistical methods to deep neural approaches, with transfer learning and meta-learning emerging as effective strategies for low-resource settings. Methods like ProtoNER and MetaNER have paved the way for few-shot NER, yet challenges remain—especially in reconciling the sequence labeling nature of NER with few-shot learning paradigms. FewTopNER builds on these advances by integrating topic modeling into the few-shot framework, offering mutual benefits for both entity recognition and topic coherence. Our work thus represents a significant step towards more robust and efficient few-shot, cross-lingual NER systems.

\section{Methodology}

\subsection{Model Architecture Overview}
\textbf{FewTopNER} introduces a novel approach to few-shot Named Entity Recognition (NER) that leverages topic-aware contextual representations to enhance entity detection across multiple languages. The model's innovation lies in its ability to incorporate document-level semantic information from topic modeling to improve few-shot NER performance, particularly in cross-lingual scenarios.
The architecture consists of four integrated components that work together to achieve this goal:
\begin{itemize}
\item A \textbf{shared encoder} based on XLM-RoBERTa that provides multilingual contextual representations enhanced by cross-lingual attention and language calibration mechanisms. This component ensures effective feature extraction across different languages while maintaining linguistic nuances \cite{Mehta2023LLM-RM}.
\item An \textbf{entity recognition branch} that implements a sophisticated prototype-based few-shot learning system. This branch combines BiLSTM encoding with a prototype network and CRF layer to enable effective entity recognition from limited examples. The prototype network learns to project entity features into a space where similar entities cluster together, facilitating few-shot learning \cite{Yan2021Unsupervised}.

\item A \textbf{topic enhancement branch} that, rather than performing traditional topic modeling, extracts and processes topic-aware features through a combination of pre-trained LDA models and neural processing. This branch projects topic features into a prototype space that complements entity recognition, focusing on enhancing NER rather than explicit topic classification \cite{Malmasi2022MultiCoNER}.

\item A \textbf{cross-task bridge} that implements bidirectional attention and feature fusion between entity and topic representations. This innovative component enables topic-aware entity recognition by allowing entity features to be enhanced by relevant topic information, while maintaining language-specific characteristics through careful gating mechanisms \cite{Schneider2022UC3M-PUCPR}.
\end{itemize}
A key distinction of FewTopNER is that it doesn't treat topic modeling as a separate classification task, but rather as an auxiliary component that enriches entity recognition. The topic branch works in concert with the entity branch through the bridge mechanism to provide additional semantic context that helps disambiguate and identify entities more accurately in few-shot scenarios \cite{Viksna2022Multilingual}. This approach is particularly effective in multilingual settings, where topic-level semantic information can help bridge linguistic differences and improve cross-lingual entity recognition performance.
The architecture's design enables three key capabilities:
\begin{itemize}
\item Effective few-shot learning through prototype-based representation learning in both entity and topic spaces
\item Enhanced entity recognition through the integration of topic-aware contextual information
\item Robust cross-lingual performance through shared representations and language-specific adaptations
\end{itemize}
\subsection{Shared Encoder}

The shared encoder represents the fundamental architectural component of FewTopNER, serving as a neural transformation mechanism that converts multilingual input sequences into semantically enriched vector representations \cite{Hedderich2020Transfer}. At its core, the encoder leverages the pre-trained XLM-RoBERTa model, which has demonstrated state-of-the-art performance in multilingual natural language processing tasks \cite{Jayanth2023Indian}. This architectural choice is motivated by XLM-RoBERTa's proven capacity to generate contextually nuanced embeddings that capture both universal linguistic patterns and language-specific nuances across diverse languages \cite{You2021Uppsala}. Through its multi-layered transformer architecture, the shared encoder processes input sequences through self-attention mechanisms and feed-forward networks, enabling the model to capture complex syntactic and semantic relationships within and across languages \cite{Xie2021PALI}. The preservation of language-specific features is achieved through specialized calibration mechanisms, while simultaneously maintaining cross-lingual alignment in the shared representation space. This dual capability is crucial for few-shot learning scenarios, where the model must effectively transfer knowledge across languages while retaining the distinctive characteristics that make each language unique.

\subsubsection{Initial Contextual Embeddings}
The initial contextual embeddings phase comprises three critical components that establish the foundation for effective multilingual representation learning. First, the input tokenization process employs XLM-RoBERTa's SentencePiece tokenizer, which implements a language-agnostic subword segmentation strategy. This tokenizer operates by decomposing input sequences into atomic subword units through a learned vocabulary of 250,000 tokens, derived from a large-scale multilingual corpus. The tokenization process preserves morphological patterns across languages while managing out-of-vocabulary words through subword decomposition, thereby ensuring robust handling of morphologically rich languages and rare tokens.
Following tokenization, the transformer layers of XLM-RoBERTa process the token embeddings through a sequence of 12 transformer blocks. Each block implements multi-head self-attention mechanisms with 12 attention heads, allowing the model to capture diverse aspects of syntactic and semantic relationships simultaneously. The self-attention operation computes attention weights through scaled dot-product attention, where each token's representation is updated based on its interactions with all other tokens in the sequence. This process is formulated as:
\[
\text{Attention}(Q, K, V) = \text{softmax}\left(\frac{QK^T}{\sqrt{d_k}}\right)V
\]
\[
\text{where } Q, K, \text{ and } V \text{ represent the query, key, and value matrices respectively, and } d_k \text{ is the dimensionality of the key vectors.}
\]
The multi-head attention mechanism projects the input into multiple subspaces, enabling the model to capture different types of dependencies in parallel:
\[
\text{MultiHead}(Q, K, V) = \text{Concat}(\text{head}_1, \ldots, \text{head}_h) W^O
\]
\[
\text{where} \quad \text{head}_i = \text{Attention}(Q W_i^Q, K W_i^K, V W_i^V)
\]

The final component generates contextual representations by aggregating information from the transformer layers. These representations, denoted as 
\(\mathbf{H} = \{ \mathbf{h}_1, \ldots, \mathbf{h}_n \}\) 
for a sequence of \(n\) tokens, encode both local syntactic patterns and global semantic dependencies. Each token representation \(\mathbf{h}_i \in \mathbb{R}^d\) maintains a dimensionality of \(d = 768\), capturing rich contextual information that serves as input for subsequent task-specific components. The contextual nature of these representations is crucial for disambiguation in named entity recognition tasks, as the same surface form may represent different entity types depending on its context.

This architectural design ensures that the initial embeddings capture both language-independent and language-specific features, providing a robust foundation for the few-shot learning objectives of FewTopNER. The representations maintain sufficient abstraction to enable cross-lingual transfer while preserving the granular features necessary for precise entity recognition.

\subsubsection{Language-Specific Calibration}
Language-specific calibration constitutes a critical component in FewTopNER's architecture, enabling effective cross-lingual transfer while preserving language-specific characteristics. The calibration process operates through four interconnected mechanisms that work in concert to achieve optimal multilingual representations.

The cross-lingual attention mechanism serves as the primary means of aligning embeddings across different languages \cite{wang2019cross}. This mechanism extends the traditional transformer attention by introducing language-aware components. For each language \(l\), we compute language-specific query (\(\mathbf{Q}_l\)), key (\(\mathbf{K}_l\)), and value (\(\mathbf{V}_l\)) matrices through learned transformations:
\[
\mathbf{Q}_l = \mathbf{W}_Q^l \mathbf{H} + \mathbf{b}_Q^l, \quad
\mathbf{K}_l = \mathbf{W}_K^l \mathbf{H} + \mathbf{b}_K^l, \quad
\mathbf{V}_l = \mathbf{W}_V^l \mathbf{H} + \mathbf{b}_V^l
\]
where \(\mathbf{H}\) represents the input hidden states, and \(\{\mathbf{W}_Q^l, \mathbf{W}_K^l, \mathbf{W}_V^l\}\) and \(\{\mathbf{b}_Q^l, \mathbf{b}_K^l, \mathbf{b}_V^l\}\) are language-specific projection matrices and bias terms, respectively. 

The attention computation incorporates language-specific biases \(\alpha_l\) to account for linguistic variations:
\[
\text{Attention}_l(\mathbf{Q}_l, \mathbf{K}_l, \mathbf{V}_l) = 
\text{softmax}\left(\frac{\mathbf{Q}_l \mathbf{K}_l^\top}{\sqrt{d_k}} + \alpha_l \right)\mathbf{V}_l
\]

The shared projection spaces are maintained through a constraint mechanism that enforces similarity between language-specific projections while allowing for controlled deviation \cite{wang2020understanding}:
\[
\mathcal{L}_\text{align} = \|\mathbf{W}_Q^l {\mathbf{W}_Q^m}^\top - \mathbf{I}\|_F + 
\|\mathbf{W}_K^l {\mathbf{W}_K^m}^\top - \mathbf{I}\|_F + 
\|\mathbf{W}_V^l {\mathbf{W}_V^m}^\top - \mathbf{I}\|_F
\]
where \(l\) and \(m\) represent different languages, and \(\|\cdot\|_F\) denotes the Frobenius norm.

To further refine the representations, we employ language-specific adaptation networks \cite{ma2021contributions}. These networks consist of feed-forward neural networks \(F_l\) that apply language-specific transformations:
\[
\mathbf{h}_l = F_l(\mathbf{h}) = \mathbf{W}_2^l \text{ReLU}(\mathbf{W}_1^l \mathbf{h} + \mathbf{b}_1^l) + \mathbf{b}_2^l
\]
where \(\mathbf{h}\) represents the input embeddings, and \(\{\mathbf{W}_1^l, \mathbf{W}_2^l, \mathbf{b}_1^l, \mathbf{b}_2^l\}\) are learned parameters specific to language \(l\). The adaptation networks are trained to optimize a combination of task-specific loss and cross-lingual alignment objectives:
\[
\mathcal{L}_\text{adapt} = \mathcal{L}_\text{task} + \lambda \mathcal{L}_\text{align}
\]

The positional encoding mechanism implements sinusoidal functions to maintain sequential information. For position \(\text{pos}\) and dimension \(i\), the encoding \(\text{PE}\) is computed as:
\[
\text{PE}(\text{pos}, 2i) = \sin\left(\frac{\text{pos}}{10000^{\frac{2i}{d_\text{model}}}}\right), \quad
\text{PE}(\text{pos}, 2i+1) = \cos\left(\frac{\text{pos}}{10000^{\frac{2i}{d_\text{model}}}}\right)
\]
This formulation ensures that relative positions are consistently encoded across different sequence lengths and languages.

Finally, the architecture incorporates residual connections and normalization layers to stabilize training and facilitate gradient flow. For each sublayer \(F\), the output is computed as:
\[
\mathbf{h}' = \text{LayerNorm}(\mathbf{h} + F(\mathbf{h}))
\]
where \(\text{LayerNorm}\) represents layer normalization:
\[
\text{LayerNorm}(\mathbf{x}) = \gamma \frac{\mathbf{x} - \mu}{\sigma} + \beta
\]
Here, \(\mu\) and \(\sigma\) are the mean and standard deviation computed over the feature dimension, while \(\gamma\) and \(\beta\) are learned scaling and shifting parameters. This normalization scheme, combined with residual connections, ensures stable gradient flow during training and enables effective learning of both language-specific and shared features.

The integration of these four mechanisms creates a robust framework for handling multilingual inputs while maintaining the balance between cross-lingual transfer and language-specific feature preservation. This calibration system proves particularly crucial in few-shot scenarios, where the model must effectively leverage knowledge across languages while respecting linguistic distinctions.

\subsubsection{Contrastive Learning for Alignment}
The contrastive learning module in FewTopNER implements a sophisticated approach to cross-lingual representation alignment, fundamentally enhancing the model's ability to transfer knowledge across languages in few-shot scenarios. This module operates on the principle that semantically equivalent content should maintain similar representations in the shared embedding space, regardless of the source language \cite{Chen2022Multi-Level}.

The alignment process employs a specialized contrastive loss function that operates on parallel sentences across different language pairs. For any given parallel sentence pair \((x_s, x_t)\) in source and target languages, their respective embeddings \((\mathbf{e}_s, \mathbf{e}_t)\) are computed through the shared encoder. The contrastive objective function is formulated as:
\[
\mathcal{L}_\text{contrast} = -\log\left(\frac{\exp(\text{sim}(\mathbf{e}_s, \mathbf{e}_t)/\tau)}{\sum_n \exp(\text{sim}(\mathbf{e}_s, \mathbf{e}_n)/\tau)}\right)
\]
where \(\text{sim}(\cdot, \cdot)\) denotes the cosine similarity between embeddings, \(\tau\) represents a temperature parameter controlling the sharpness of the distribution, and \(\mathbf{e}_n\) includes both the positive target embedding and a set of negative samples drawn from other sentences in the batch. This formulation encourages the model to minimize the distance between parallel content while maintaining discriminative power for distinct semantic concepts.

To preserve language-specific characteristics while promoting cross-lingual alignment, we introduce a regularization term that balances these competing objectives:
\[
\mathcal{L}_\text{reg} = \mathcal{L}_\text{contrast} + \lambda \mathcal{L}_\text{diversity}
\]
The diversity loss \(\mathcal{L}_\text{diversity}\) penalizes excessive homogenization of embeddings, ensuring that important linguistic nuances are maintained:
\[
\mathcal{L}_\text{diversity} = \max\left(0, \mu - \frac{1}{N} \sum_{i,j} \|\mathbf{e}_i - \mathbf{e}_j\|_2\right)
\]
where \(\mu\) represents a margin hyperparameter, and the summation operates over all pairs of embeddings within a mini-batch. This term ensures that embeddings maintain sufficient distinctiveness to encode language-specific features while still achieving cross-lingual alignment.

The effectiveness of this contrastive learning approach is further enhanced through curriculum learning, where the complexity of alignment tasks gradually increases during training. Initially, the model focuses on aligning highly similar parallel sentences, progressively advancing to more challenging cases involving idiomatic expressions and culture-specific references. This curriculum is implemented through a dynamic sampling strategy \cite{Xiao2024LACNER}:
\[
p(x_s, x_t) \propto \exp(-\beta \cdot d(x_s, x_t))
\]
where \(d(\cdot, \cdot)\) measures the semantic distance between parallel sentences, and \(\beta\) controls the sampling temperature throughout training.

The integration of this contrastive learning module with the broader FewTopNER architecture creates a robust framework for cross-lingual representation learning. The resulting embeddings demonstrate strong performance in few-shot scenarios, effectively leveraging knowledge across languages while maintaining the ability to capture language-specific nuances essential for accurate named entity recognition.

\subsection{Entity Recognition Branch}

The entity recognition branch constitutes a crucial component of FewTopNER, implementing an innovative prototype-based learning framework specifically designed to address the challenges inherent in low-resource named entity recognition scenarios. This architectural component advances beyond traditional sequence labeling approaches by introducing a meta-learning strategy that enables effective entity detection and classification with minimal annotated examples. Through the integration of prototype networks with sophisticated feature extraction mechanisms, the branch learns to construct and maintain discriminative entity representations that generalize effectively across different entity types and linguistic contexts.
At its theoretical foundation, the branch leverages the concept of prototypical networks, extending their application to the sequential nature of named entity recognition tasks. Unlike conventional approaches that require extensive labeled data to learn entity patterns, this framework operates on the principle of learning to learn from few examples. It accomplishes this by maintaining a dynamic prototype space where entity representations are continuously refined through episodic training \cite{Moscato2023Few-shot}, allowing the model to capture essential entity characteristics while minimizing the dependency on large annotated datasets.
The design of this branch is motivated by two key observations in low-resource NER: first, that entity types often share common structural and contextual patterns across languages, and second, that effective few-shot learning requires the ability to rapidly adapt to new entity categories while maintaining stability on previously learned ones. These insights inform the implementation of a hierarchical feature extraction pipeline that combines local contextual information with global semantic understanding, enabling robust entity recognition even in scenarios where labeled data is scarce.

\subsubsection{Entity Encoder}
The Entity Encoder implements a sophisticated three-tier architecture designed to process and enhance token representations for effective named entity recognition. This component builds upon the initial embeddings provided by the shared encoder, introducing specialized layers that capture hierarchical linguistic patterns essential for entity detection and classification.

The first tier employs a Bidirectional Long Short-Term Memory (BiLSTM) network to model sequential dependencies. For a given input sequence \(\mathbf{X} = \{ x_1, x_2, \dots, x_n \}\), the BiLSTM processes the tokens in both forward and backward directions, computing:
\[
h_t^f = \text{LSTM}_f(x_t, h_{t-1}^f), \quad
h_t^b = \text{LSTM}_b(x_t, h_{t+1}^b)
\]
The concatenated output \(\mathbf{h}_t = [h_t^f; h_t^b]\) captures both preceding and following context for each token, enabling the model to understand entity boundaries and internal structure. This bidirectional processing is particularly crucial for identifying entity spans that depend on both left and right context, such as person names or organizational titles.

The second tier introduces language-specific adapters that fine-tune the representations for each supported language. These adapters implement a bottleneck architecture:
\[
\mathbf{h}_l = \mathbf{W}_2^l \text{ReLU}(\mathbf{W}_1^l \mathbf{h} + \mathbf{b}_1^l) + \mathbf{b}_2^l
\]
where \(\{\mathbf{W}_1^l, \mathbf{W}_2^l\}\) and \(\{\mathbf{b}_1^l, \mathbf{b}_2^l\}\) are language-specific parameters learned during training. The bottleneck design, with \(\mathbf{W}_1^l \in \mathbb{R}^{d \times r}\) and \(\mathbf{W}_2^l \in \mathbb{R}^{r \times d}\), where \(r < d\), enforces efficient parameter usage while maintaining language-specific customization. This adaptation mechanism allows the model to account for language-specific entity formation patterns and contextual cues.

The third tier implements multi-head self-attention to aggregate token-level dependencies, computing attention scores:
\[
\text{Attention}(\mathbf{Q}, \mathbf{K}, \mathbf{V}) = \text{softmax}\left(\frac{\mathbf{Q}\mathbf{K}^\top}{\sqrt{d_k}}\right)\mathbf{V}
\]
with multiple attention heads operating in parallel:
\[
\text{MultiHead}(\mathbf{H}) = \text{Concat}(\text{head}_1, \dots, \text{head}_h)\mathbf{W}^O
\]
where \(\text{head}_i = \text{Attention}(\mathbf{H}\mathbf{W}_i^Q, \mathbf{H}\mathbf{W}_i^K, \mathbf{H}\mathbf{W}_i^V)\). This multi-head attention mechanism enables the model to capture different types of dependencies simultaneously, with each head potentially specializing in specific aspects of entity-related patterns. The resulting contextual representations incorporate both local sequential information from the BiLSTM and global dependencies from the attention mechanism.

The complete entity encoding process can be expressed as:
\[
\mathbf{E}(\mathbf{X}) = \text{MultiHead}(\text{Adapt}_l(\text{BiLSTM}(\mathbf{X})))
\]
where \(\mathbf{E}(\mathbf{X})\) represents the final entity-aware representations that serve as input to the subsequent prototype learning stages. This hierarchical processing ensures that the encoder captures both the sequential nature of entity spans and the complex interdependencies between tokens, while maintaining language-specific adaptability crucial for cross-lingual scenarios.

\subsubsection{Prototype Network}
The Prototype Network constitutes a critical component of FewTopNER's entity recognition branch, implementing a metric-based learning approach that enables effective few-shot adaptation to new entity types. This component operates on the principle that entity classes can be represented by prototypical vectors in a learned metric space, where classification decisions are made based on distances to these prototypes \cite{Fritzler2018Few-shot} .

For each entity type \( e \) in the set of target entity classes \( \mathcal{E} \), the prototype computation process aggregates information from support set examples \( S_e = \{ (x_1, y_1), \dots, (x_k, y_k) \} \) through a weighted averaging mechanism. The prototype \( p_e \) for entity type \( e \) is computed as:
\[
p_e = \frac{\sum_i \alpha_i f_\theta(x_i)}{\sum_i \alpha_i}
\]
where \( f_\theta \) represents the entity encoder function parameterized by \( \theta \), and \( \alpha_i \) are attention-based weights that determine the contribution of each support example. These weights are computed through a similarity-based attention mechanism:
\[
\alpha_i = \text{softmax}(f_\theta(x_i)^T \mathbf{W}_a f_\theta(x_q))
\]
where \( x_q \) represents the query token being classified, and \( \mathbf{W}_a \) is a learnable attention matrix. This attention-weighted prototype computation ensures that the model can effectively handle noisy or ambiguous support examples while capturing the most representative features of each entity type.

The distance-based classification mechanism operates in the metric space induced by the entity encoder. For a query token \( x_q \), the probability distribution over entity types is computed using a softmax over negative distances:
\[
P(y = e | x_q) = \frac{\exp(-d(f_\theta(x_q), p_e))}{\sum_{e'} \exp(-d(f_\theta(x_q), p_{e'}))}
\]
where \( d(\cdot, \cdot) \) represents the distance metric, typically implemented as Euclidean distance or cosine similarity. This formulation ensures that the probability assignment is inversely proportional to the distance between the query embedding and each entity prototype.

The prototype memory bank introduces an efficient mechanism for maintaining and updating prototypes during both training and inference \cite{Ji2022Few-shot}. The memory bank \( M \) maintains a dynamic set of prototypes \( \{p_e\}_{e \in \mathcal{E}} \) along with their associated statistics:
\[
M = \{(p_e, n_e, \sigma_e)\}_{e \in \mathcal{E}}
\]
where \( n_e \) represents the number of examples seen for entity type \( e \), and \( \sigma_e \) captures the variance of features around the prototype. The update mechanism for the memory bank follows an exponential moving average:
\[
p_e^{\text{new}} = (1 - \gamma) p_e + \gamma p_e^{\text{batch}}, \quad \sigma_e^{\text{new}} = (1 - \gamma) \sigma_e + \gamma \sigma_e^{\text{batch}}
\]
where \( \gamma \) is a momentum coefficient that controls the rate of prototype adaptation, and \( p_e^{\text{batch}} \) and \( \sigma_e^{\text{batch}} \) are computed from the current batch of examples. This momentum-based updating ensures stable prototype evolution while allowing adaptation to new examples.

The integration of these three components—prototype computation, distance-based classification, and the prototype memory bank—creates a robust framework for few-shot entity recognition. The system can rapidly adapt to new entity types while maintaining stable performance on previously learned categories, making it particularly effective in low-resource scenarios where labeled data is scarce.

\subsubsection{Conditional Random Fields (CRF)}
The Conditional Random Fields layer represents the final component of the entity recognition branch, implementing a structured prediction framework that enforces coherent entity labeling across entire sequences. Unlike token-level classification approaches that make independent decisions for each token, the CRF layer explicitly models the interdependencies between adjacent entity tags, capturing crucial sequential patterns in entity formation.

The CRF layer operates by defining a conditional probability distribution over the entire sequence of tags \( Y = \{y_1, y_2, \dots, y_n\} \) given the input sequence \( X = \{x_1, x_2, \dots, x_n\} \):
\[
P(Y | X) = \frac{\exp(\text{score}(X,Y))}{Z(X)}
\]
where \( \text{score}(X,Y) \) combines both emission and transition scores:
\[
\text{score}(X,Y) = \sum_i \left( E(y_i | x_i) + T(y_{i-1}, y_i) \right)
\]
Here, \( E(y_i | x_i) \) represents emission scores from the prototype network, while \( T(y_{i-1}, y_i) \) captures transition scores between adjacent tags. The normalization term \( Z(X) \) sums over all possible tag sequences:
\[
Z(X) = \sum_{\mathbf{y}} \exp(\text{score}(X, \mathbf{y}))
\]
The transition matrix \( T \) learns valid entity tag transitions, encoding constraints such as:
\begin{itemize}
    \item I-PER tags must follow B-PER or I-PER tags.
    \item B-ORG cannot directly follow I-PER.
    \item O tags can transition to any B- tag but not to I- tags.
\end{itemize}

During inference, the Viterbi algorithm efficiently computes the optimal tag sequence \( Y^* \) that maximizes the conditional probability:
\[
Y^* = \arg\max_Y P(Y | X)
\]
This dynamic programming approach recursively computes the maximum score for each position \( t \) and tag \( k \):
\[
V_{t,k} = \max_j \left( V_{t-1,j} + T_{j,k} + E_{k,t} \right)
\]
where \( V_{t,k} \) represents the maximum score ending at position \( t \) with tag \( k \). The algorithm maintains backpointers to reconstruct the optimal path once the forward pass completes.

The integration of the CRF layer with the prototype network creates a powerful synergy: while the prototype network excels at learning discriminative entity representations from few examples, the CRF layer ensures that the predicted entity spans maintain structural consistency across the sequence. This combination proves particularly valuable in handling complex entity patterns that span multiple tokens or involve nested structures.

\subsection{Topic Modeling Branch}

The topic modeling branch represents a novel fusion of classical statistical approaches and modern neural architectures, designed to extract and leverage document-level semantic information for enhanced named entity recognition. This component advances beyond traditional topic modeling frameworks by integrating both probabilistic topic inference and neural representation learning, creating a synergistic system that captures semantic relationships at multiple levels of granularity.
At its theoretical foundation, the branch builds upon the insight that entity recognition can be significantly enhanced by understanding the broader thematic context in which entities appear. For instance, documents discussing technological innovations are more likely to contain company names and technical terms, while news articles about sports events frequently mention athlete names and team organizations. This contextual awareness becomes particularly crucial in few-shot scenarios, where limited training examples make it essential to leverage every available signal for accurate entity classification \cite{Doan2021Benchmarking}.
The architecture implements a hybrid approach that combines the interpretability and statistical rigor of traditional topic models with the representational power of neural networks. By processing document collections through both probabilistic topic inference and neural encoding pathways, the branch constructs rich topic representations that capture both explicit thematic patterns and implicit semantic relationships. These topic representations are then carefully integrated with token-level features, enabling the model to make entity recognition decisions that are informed by both local syntactic patterns and global thematic context.

\subsubsection{Topic Encoder}
The Topic Encoder implements a sophisticated multi-stage architecture that combines classical probabilistic topic modeling with modern neural approaches to create rich semantic representations. This hybrid design addresses the fundamental challenge of capturing both explicit thematic patterns and latent semantic relationships in multilingual documents.

For the Language-Specific LDA stage, we employ separate Latent Dirichlet Allocation models trained on monolingual corpora for each supported language \( l \). The probability of a document \( d \) generating a word \( w \) is formulated as:
\[
P(w|d,l) = \sum_k P(w|z_k,l) P(z_k|d)
\]
where \( z_k \) represents the \( k \)-th topic, and the language-specific word distributions \( P(w|z_k,l) \) capture vocabulary patterns unique to each language. The document-topic distributions \( \theta_d = P(z|d) \) are inferred using collapsed Gibbs sampling, providing interpretable initial topic representations. These distributions are computed as:
\[
\theta_d = \frac{n_d + \alpha}{\sum_k n_{d,k} + K\alpha}
\]
where \( n_{d,k} \) represents the count of words assigned to topic \( k \) in document \( d \), \( K \) is the total number of topics, and \( \alpha \) is the Dirichlet prior parameter.

The Neural Feature Fusion mechanism enriches these statistical topic representations by combining them with contextual embeddings. For each document \( d \), we compute a fused representation \( f_d \):
\[
f_d = \sigma(W_f [\theta_d; h_d] + b_f)
\]
where \( h_d \) represents the document's neural embedding obtained from the shared encoder, \( [\;] \) denotes concatenation, and \( \{W_f, b_f\} \) are learnable parameters. The activation function \( \sigma \) (typically GeLU) introduces non-linearity, allowing the model to capture complex interactions between statistical and neural features.

The Adaptive Pooling stage implements a dynamic aggregation mechanism that adjusts to document structure. For a document with \( n \) tokens, the pooled representation \( p_d \) is computed through an attention-weighted sum:
\[
p_d = \sum_i \alpha_i h_i
\]
where the attention weights \( \alpha_i \) are computed as:
\[
\alpha_i = \text{softmax}(v^T \tanh(W_p h_i + W_t f_d))
\]
Here, \( \{W_p, W_t, v\} \) are learnable parameters, and the attention mechanism allows the model to focus on tokens that are most relevant to the document's topical content. The resulting representation combines both token-level and document-level semantic information.

This three-stage process creates topic representations that are both interpretable through the LDA component and richly expressive through neural enhancement. The final output serves as a crucial input to subsequent components of FewTopNER, providing semantic context that aids in entity disambiguation and classification.

\subsubsection{Topic Prototype Network}
The Topic Prototype Network extends the principles of metric-based few-shot learning to the domain of topic modeling, establishing a framework that allows rapid adaptation to new topics while maintaining semantic coherence  \cite{Iwata2021Few-shot}. This component operates in a specialized metric space where topic representations can be effectively compared and classified.

The prototype representation mechanism implements a non-linear projection function that maps topic features into a shared embedding space where semantic relationships are preserved. For a given topic representation \( t \), the projection is computed through a series of transformations:
\[
\phi(t) = W_2 \, \text{ReLU}(W_1 t + b_1) + b_2
\]
where \( \{W_1, W_2, b_1, b_2\} \) are learnable parameters. This projected representation ensures that topics with similar semantic content cluster together in the embedding space. The prototype for each topic category \( c \) is then computed as a weighted average of support set examples:
\[
p_c = \frac{\sum_i \alpha_i \phi(t_i)}{\sum_i \alpha_i}
\]
The attention weights \( \alpha_i \) are determined through a learned similarity function that considers both feature similarity and topic coherence:
\[
\alpha_i = \text{softmax}(s(\phi(t_i), \phi(t_q)) + \lambda c(t_i))
\]
where \( s(\cdot, \cdot) \) measures feature similarity, \( t_q \) represents the query topic, \( c(\cdot) \) evaluates topic coherence, and \( \lambda \) balances these two factors.

The similarity metrics component employs a temperature-scaled cosine similarity function to compute classification probabilities. For a query topic \( t_q \), the probability of belonging to topic category \( c \) is:
\[
P(c|t_q) = \frac{\exp(\cos(\phi(t_q), p_c) / \tau)}{\sum_{c'} \exp(\cos(\phi(t_q), p_{c'}) / \tau)}
\]
where \( \tau \) is a learnable temperature parameter that controls the sharpness of the probability distribution. This formulation ensures that the model can make confident predictions when appropriate while maintaining uncertainty in ambiguous cases.

To preserve topic coherence throughout the learning process, the network incorporates several regularization mechanisms. The primary coherence loss term encourages topics to maintain interpretable word distributions:
\[
L_\text{coherence} = -\sum_c \sum_{w,w'} \text{PMI}(w, w') P(w|c) P(w'|c)
\]
where \( \text{PMI}(w, w') \) represents the pointwise mutual information between words \( w \) and \( w' \), and \( P(w|c) \) is the probability of word \( w \) under topic \( c \). Additional regularization terms prevent prototype collapse and maintain diversity:
\[
L_\text{diversity} = -\sum_{c \neq c'} ||p_c - p_{c'}||_2
\]

The complete loss function combines these objectives with the classification loss:
\[
L_\text{total} = L_\text{class} + \alpha_1 L_\text{coherence} + \alpha_2 L_\text{diversity}
\]
where \( \alpha_1 \) and \( \alpha_2 \) are hyperparameters controlling the relative importance of each term.

This sophisticated combination of prototype learning and coherence preservation enables the Topic Prototype Network to effectively capture and classify semantic patterns while maintaining interpretable topic representations. The resulting topic prototypes serve as reliable reference points for enhancing entity recognition through semantic context.

\subsection{Cross-Task Bridge}

The Cross-Task Bridge represents a sophisticated architectural component that facilitates dynamic information exchange between named entity recognition and topic modeling processes. This bidirectional communication channel enables each branch to leverage complementary information from the other, enhancing both entity detection and topic understanding simultaneously.

\subsubsection{Cross-Task Attention}

The task-specific projection mechanism transforms features from both branches into compatible representation spaces while preserving their essential characteristics. For entity representations \( E \in \mathbb{R}^{d_e} \) and topic representations \( T \in \mathbb{R}^{d_t} \), the projections are computed through learnable transformations:
\[
E' = W_E E + b_E
\]
\[
T' = W_T T + b_T
\]
where \( W_E, W_T \in \mathbb{R}^{d_s \times d_e} \) and \( b_E, b_T \in \mathbb{R}^{d_s} \) project both feature sets into a shared dimensionality \( d_s \). These projections are designed to maintain task-specific information while enabling meaningful cross-task comparisons.

The multi-head attention mechanism implements a sophisticated cross-task information exchange through multiple attention heads, each capturing different aspects of the relationship between entities and topics. For head \( h \), the attention computation is formulated as:
\[
A_h(E', T') = \text{softmax}\left(\frac{(E'W_Q^h)(T'W_K^h)^\top}{\sqrt{d_k}}\right)T'W_V^h
\]
where \( W_Q^h, W_K^h, W_V^h \) are head-specific projection matrices. The multi-head output is computed as:
\[
\text{MultiHead}(E', T') = \text{Concat}(A_1, \ldots, A_H)W_O
\]
This mechanism allows each head to focus on different semantic relationships between entities and topics, such as contextual relevance, semantic similarity, or structural patterns.

The language awareness component introduces dynamic attention scaling based on language-specific features. For a given language \( l \), the attention weights are adjusted through a language-dependent scaling factor:
\[
\alpha_l = \sigma(W_l[h_l; g_l] + b_l)
\]
where \( h_l \) represents language-specific hidden states, \( g_l \) captures global language characteristics, and \( \sigma \) denotes a sigmoid activation. The final attention computation incorporates these language-aware scaling factors:
\[
\text{Attention}_l(E', T') = \alpha_l \odot \text{MultiHead}(E', T')
\]
where \( \odot \) represents element-wise multiplication. This language-aware scaling ensures that the cross-task attention mechanism appropriately considers linguistic variations and patterns specific to each language \cite{Zhu2022Dual}.

The complete cross-task attention output provides enriched representations for both tasks:
\[
E_\text{enriched} = \text{LayerNorm}(E + \text{Attention}_l(E', T'))
\]
\[
T_\text{enriched} = \text{LayerNorm}(T + \text{Attention}_l(T', E'))
\]
These enriched representations enable more accurate entity recognition by incorporating topic-level semantic context, while simultaneously refining topic understanding through entity-level information.

\subsubsection{Task Gating System}
The Task Gating System implements an adaptive fusion mechanism that selectively combines features from both tasks while maintaining their distinct characteristics. This sophisticated gating architecture ensures that information exchange between tasks occurs only when it provides meaningful benefits to either branch's performance.

At its core, the gating mechanism computes dynamic fusion weights through a learned function that evaluates the relevance of cross-task features. For entity features \( e \) and topic features \( t \), the gating operation is formulated as:
\[
g = \sigma(W_g [e; t] + b_g)
\]
where \( [e; t] \) represents the concatenation of entity and topic features, \( \{W_g, b_g\} \) are learnable parameters, and \( \sigma \) denotes the sigmoid activation function. The resulting gate values \( g \in [0,1] \) determine the degree of feature mixing between tasks.

The feature fusion process incorporates these gates through a weighted combination:
\[
f_\text{fused} = g \odot (W_e e + W_t t) + (1-g) \odot f_\text{original}
\]
where \( \{W_e, W_t\} \) are task-specific projection matrices, \( \odot \) represents element-wise multiplication, and \( f_\text{original} \) denotes the original features from each respective branch. This formulation allows the model to maintain task-specific processing when cross-task information is not beneficial while enabling rich feature integration when complementary information is available.

To ensure stable training and preserve task-specific knowledge, the system employs carefully designed residual connections. These connections are implemented through a hierarchical structure:
\[
h^l = \text{LayerNorm}(h^{l-1} + F^l(h^{l-1}))
\]
where \( F^l \) represents the transformation at layer \( l \), and \(\text{LayerNorm}\) denotes layer normalization. The residual paths serve multiple crucial purposes:
\begin{itemize}
    \item They provide direct gradient flow to earlier layers, facilitating stable optimization.
    \item They help preserve task-specific features by maintaining a direct path for original information.
    \item They mitigate the risk of catastrophic forgetting by ensuring that fundamental task knowledge remains accessible.
\end{itemize}

The complete gating system integrates these components to produce the final output:
\[
\text{output} = \alpha_\text{residual} h_\text{original} + (1-\alpha_\text{residual}) h_\text{gated}
\]
where \( \alpha_\text{residual} \) is a learned parameter that balances the contribution of original and gated features. This careful orchestration of feature fusion and preservation enables the model to leverage cross-task information effectively while maintaining robust performance on each individual task.

\subsection{Training Framework}

FewTopNER employs a multi-faceted training strategy to balance task-specific and cross-task objectives.

\subsubsection{Episode Construction}
The Episode Construction mechanism implements a sophisticated sampling strategy that enables effective few-shot learning across multiple languages and tasks. This component systematically organizes training data into episodic structures that simulate real-world scenarios where only limited labeled data is available for new entity types and topics.

The \textbf{N-Way K-Shot} sampling process creates balanced training episodes by carefully selecting examples from the available data. For each episode, the framework samples \( N \) distinct entity types and selects \( K \) examples for each type, forming a structured learning environment. This sampling is formulated mathematically as:
\[
E = \{(x_i, y_i, l_i)\}_{i=1}^{N \times K}
\]
where \( x_i \) represents an input sequence, \( y_i \) denotes its corresponding entity labels, and \( l_i \) indicates the language. The sampling probability for each entity type \( e \) is computed through a temperature-scaled distribution:
\[
P(e) = \frac{\exp(f(e)/\tau)}{\sum_{e'} \exp(f(e')/\tau)}
\]
where \( f(e) \) represents the frequency of entity type \( e \) in the training data, and \( \tau \) is a temperature parameter that controls sampling uniformity. Lower values of \( \tau \) promote more uniform sampling across entity types, while higher values maintain the natural distribution.

\textbf{Cross-lingual episode construction} extends this sampling framework to multiple languages simultaneously. For each episode, the framework ensures representation from different language pairs while maintaining semantic alignment. The cross-lingual sampling strategy is defined as:
\[
P(l_1, l_2) = \frac{\exp(s(l_1, l_2)/\gamma)}{\sum_{l'_1, l'_2} \exp(s(l'_1, l'_2)/\gamma)}
\]
where \( s(l_1, l_2) \) measures the semantic similarity between languages \( l_1 \) and \( l_2 \), and \( \gamma \) controls the degree of cross-lingual mixing. This approach ensures that episodes contain complementary examples across languages while maintaining semantic coherence.

The \textbf{support and query set division} implements a strategic split of the sampled data within each episode:
\[
S = \{(x_i, y_i, l_i)\}_{i=1}^{N \times K_s} \quad \text{(Support set)}
\]
\[
Q = \{(x_j, y_j, l_j)\}_{j=1}^{N \times K_q} \quad \text{(Query set)}
\]
where \( K_s \) and \( K_q \) represent the number of examples per class in the support and query sets, respectively. This division simulates the real-world scenario where a model must learn from limited examples (\textit{support set}) and generalize to unseen instances (\textit{query set}). The relationship between these sets is carefully managed to maintain:
\begin{itemize}
    \item \textbf{Balanced representation} of entity types and topics
    \item \textbf{Consistent distribution} of linguistic patterns
    \item \textbf{Appropriate difficulty scaling} for meta-learning
\end{itemize}

The entire episode construction process can be represented as a function:
\[
E = \text{ConstructEpisode}(D, N, K_s, K_q, \{l_1, \dots, l_m\})
\]
where \( D \) represents the complete training dataset, and \( \{l_1, \dots, l_m\} \) are the target languages. This function ensures that each constructed episode provides a meaningful learning opportunity while maintaining the constraints necessary for effective few-shot learning.

\subsubsection{Loss Computation}
The loss computation system implements a carefully balanced multi-objective optimization framework that guides the model's learning across its interconnected components. This sophisticated loss architecture combines five distinct components, each targeting specific aspects of the model's performance while working together to achieve optimal overall behavior.
\begin{itemize}
\item \textbf{Named Entity Recognition (NER) Loss: }
The NER Loss combines two complementary terms to ensure accurate entity recognition:
\[
L_{\text{NER}} = -\sum_i \log P(y_i|x_i) + \lambda_{\text{CRF}} \log P(Y|X)
\]
where the first term represents the token-level cross-entropy loss for each token \( i \), and the second term captures sequence-level consistency through the CRF scoring function. The parameter \( \lambda_{\text{CRF}} \) balances these two objectives. The CRF component specifically computes:
\[
\log P(Y|X) = \text{score}(X, Y) - \log\left(\sum_{y} \exp(\text{score}(X, y))\right)
\]
where \( \text{score}(X, Y) \) combines emission and transition scores for the entire sequence.

\item \textbf{Topic Coherence Loss: }
The Topic Coherence Loss ensures that extracted topics maintain semantic meaningfulness:
\[
L_{\text{topic}} = -\sum_k \sum_{i,j} \text{PMI}(w_i, w_j) P(w_i|z_k)P(w_j|z_k) + \lambda_{\text{div}} \|PP^\top - I\|_F^2
\]
The first term maximizes the pointwise mutual information (PMI) between frequently co-occurring words within topics, while the second term encourages diversity among topic prototypes \( P \) through orthogonality constraints. The parameter \( \lambda_{\text{div}} \) controls the strength of the diversity constraint.

\item \textbf{Cross-Task Alignment Loss: }
The Cross-Task Alignment Loss promotes effective information exchange between the NER and topic modeling branches:
\[
L_{\text{align}} = \|F_E(H_E) - F_T(H_T)\|^2 + \lambda_{\text{mutual}} I(H_E; H_T)
\]
where \( F_E \) and \( F_T \) represent feature projections from each branch, \( H_E \) and \( H_T \) are the respective hidden representations, and \( I(\cdot;\cdot) \) denotes mutual information. This loss ensures that the branches learn complementary but aligned representations.

\item \textbf{Contrastive Loss: }
The Contrastive Loss enhances cross-lingual representation learning \cite{Han2021Cross-Level}:
\[
L_{\text{contrast}} = -\log\left(\frac{\exp(s(h_a, h_p)/\tau)}{\sum_k \exp(s(h_a, h_k)/\tau)}\right)
\]
where \( h_a \) and \( h_p \) are anchor and positive example embeddings from different languages, \( s(\cdot, \cdot) \) is cosine similarity, \( \tau \) is a temperature parameter, and the summation in the denominator runs over all negative examples in the batch.

\item \textbf{Regularization Term: }
The Regularization term implements entropy-based constraints to prevent overfitting:
\[
L_{\text{reg}} = -H(P_E) - H(P_T) + \lambda_{\text{smooth}} \|\theta\|^2
\]
where \( H(\cdot) \) represents the Shannon entropy of the predicted distributions for entities (\( P_E \)) and topics (\( P_T \)), and the \( \ell_2 \)-norm of model parameters \( \theta \) provides additional smoothing.

\item \textbf{Total Loss Function: }
The complete loss function combines these components with dynamically learned weights:
\[
L_{\text{total}} = w_{\text{NER}} L_{\text{NER}} + w_{\text{topic}} L_{\text{topic}} + w_{\text{align}} L_{\text{align}} + w_{\text{contrast}} L_{\text{contrast}} + w_{\text{reg}} L_{\text{reg}}
\]
where \( \{w_{\text{NER}}, w_{\text{topic}}, w_{\text{align}}, w_{\text{contrast}}, w_{\text{reg}}\} \) are dynamically adjusted during training through a meta-learning process that optimizes their values based on validation performance.

\end{itemize}

\subsubsection{Optimization Strategy}

The optimization strategy implements a carefully orchestrated approach to training FewTopNER's interconnected components, ensuring stable convergence while maintaining the delicate balance between multiple learning objectives. The strategy combines advanced optimization techniques with specialized adaptations for multilingual few-shot learning.

\begin{itemize}
\item \textbf{AdamW Optimizer: }

The AdamW optimizer serves as the foundation of the training process, implementing a weight-decay-corrected variant of adaptive moment estimation. For each parameter \( \theta \) in module \( m \), the update rule is defined as:
\[
\hat{m}_t = \beta_1 m_{t-1} + (1-\beta_1) g_t
\]
\[
\hat{v}_t = \beta_2 v_{t-1} + (1-\beta_2) g_t^2
\]
\[
\theta_t = \theta_{t-1} - \eta_m \frac{\hat{m}_t}{\sqrt{\hat{v}_t} + \epsilon} - \lambda_m \theta_{t-1}
\]
where:
\begin{itemize}
    \item \( \eta_m \): Module-specific learning rate
    \item \( \lambda_m \): Weight decay factor
    \item \( \beta_1, \beta_2 \): Momentum coefficients
    \item \( \epsilon \): Numerical stability constant
\end{itemize}

The learning rates for each architectural component are carefully tuned:
\[
\begin{aligned}
    \text{Shared Encoder:} & \quad \eta_{\text{se}} = 1 \times 10^{-5} \\
    \text{Entity Recognition Branch:} & \quad \eta_{\text{er}} = 2 \times 10^{-5} \\
    \text{Topic Modeling Branch:} & \quad \eta_{\text{tm}} = 3 \times 10^{-5} \\
    \text{Cross-Task Bridge:} & \quad \eta_{\text{cb}} = 2 \times 10^{-5}
\end{aligned}
\]

\item \textbf{Learning Rate Scheduling: }

The learning rate scheduling follows the following pattern:
\begin{itemize}
    \item \textbf{Warmup Phase (0 \( \leq \) t \( \leq \) t\(_{\text{warmup}}\)):} The learning rate increases linearly:
    \[
    \eta(t) = \eta_{\text{base}} \cdot \frac{t}{t_{\text{warmup}}}
    \]
\end{itemize}

\item \textbf{Gradient Clipping: }

Gradient clipping prevents explosive gradients by constraining their global norm:
\[
g_{\text{clipped}} = g \cdot \min\left(1, \frac{\gamma}{\|g\|_2}\right)
\]
where \( \gamma \) is the clipping threshold (typically set to \( 1.0 \)).

\item \textbf{Language-Specific Gradient Scaling: }

To balance multilingual learning, a dynamic scaling mechanism adjusts the contribution of each language. For language \( l \), the scaling factor \( \alpha_l \) is computed as:
\[
\alpha_l = \text{softmax}(w_l \cdot h_l)
\]
where:
\begin{itemize}
    \item \( w_l \): Learnable language-specific weights
    \item \( h_l \): Language statistics (e.g., vocabulary size, syntactic complexity)
\end{itemize}

The final gradients for each language are scaled as:
\[
g_{\text{final}} = \alpha_l \cdot g_{\text{clipped}}
\]

\item \textbf{Training Routine: }

The complete optimization process integrates these components through the following training routine:
\begin{enumerate}
    \item \textbf{Initialization:} Initialize all components with pretrained weights where available.
    \item \textbf{Warmup Phase:} Apply the warmup schedule with gradient accumulation for stable initialization.
    \item \textbf{Main Training Phase:} Transition to the main training phase using cosine decay scheduling.
    \item \textbf{Monitoring:} Monitor validation metrics for each language to adjust scaling factors \( \alpha_l \).
    \item \textbf{Early Stopping:} Apply early stopping based on cross-lingual performance plateaus.
\end{enumerate}

\end{itemize}

\subsection{Data Processing Pipeline}

\subsubsection{Dataset Management}

The Dataset Management system implements a sophisticated data handling framework to ensure consistent and efficient processing of multilingual information for named entity recognition (NER) and topic modeling tasks. This system creates unified representations to effectively capture both entity-level and topic-level features while maintaining cross-lingual compatibility \cite{Yang2022CROP} .

\begin{itemize} 
\item \textbf{Unified Representation Creation}

The system uses a hierarchical feature processing approach to create standardized representations. For each input sequence \( X \), the unified feature representation is defined as:
\[
F(X) = [f_{\text{token}}; f_{\text{subword}}; f_{\text{contextual}}]
\]
where:
\begin{itemize}
    \item \( f_{\text{token}} \): Basic token features
    \item \( f_{\text{subword}} \): Morphological information obtained through subword tokenization
    \item \( f_{\text{contextual}} \): Broader document context features
\end{itemize}

To ensure consistent scaling across languages and feature types, the features are normalized as:
\[
F_{\text{norm}}(X) = \frac{F(X) - \mu_l}{\sigma_l}
\]
where:
\begin{itemize}
    \item \( \mu_l \): Language-specific mean vector
    \item \( \sigma_l \): Language-specific standard deviation vector
\end{itemize}
These statistics (\( \mu_l, \sigma_l \)) are computed from the training data for each language \( l \).

\item \textbf{Cross-Lingual Alignment: }

To ensure cross-lingual compatibility, the system computes an alignment matrix \( A_{l_1, l_2} \) for each language pair \( (l_1, l_2) \). The alignment matrix is optimized as:
\[
A_{l_1, l_2} = \operatorname*{argmin}_A \| X_{l_1}A - X_{l_2} \|^2 + \lambda \| A \|_1
\]
where:
\begin{itemize}
    \item \( X_{l_1} \): Features from language \( l_1 \)
    \item \( X_{l_2} \): Features from language \( l_2 \)
    \item \( \| A \|_1 \): Sparsity constraint to promote meaningful mappings
\end{itemize}
This optimization balances feature alignment accuracy with computational efficiency.

\item \textbf{Dynamic Batching: }

Dynamic batching improves computational efficiency by adapting batch sizes during training. Each batch is constructed as:
\[
B_t = \text{ConstructBatch}(D, s_t, \{l_1, ..., l_m\})
\]
where:
\begin{itemize}
    \item \( D \): Dataset
    \item \( s_t \): Current sampling strategy, evolving based on performance metrics
    \item \( \{l_1, ..., l_m\} \): Target languages
\end{itemize}

The batch size \( b_t \) is dynamically adjusted as:
\[
b_t = \min(b_{\text{max}}, \lfloor c \cdot \sqrt{|V_l|} \rfloor)
\]
where:
\begin{itemize}
    \item \( |V_l| \): Vocabulary size for language \( l \)
    \item \( c \): Scaling constant
    \item \( b_{\text{max}} \): Maximum batch size
\end{itemize}

\item \textbf{Feature Normalization: }

Feature normalization is implemented layer-wise, accounting for both global and language-specific statistics. The normalized feature representation is computed as:
\[
x_{\text{norm}} = \gamma_l \cdot \frac{x - \mu_g}{\sqrt{\sigma_g^2 + \epsilon}} + \beta_l
\]
where:
\begin{itemize}
    \item \( \mu_g, \sigma_g \): Global mean and standard deviation
    \item \( \gamma_l, \beta_l \): Language-specific scaling and shifting parameters (learned during training)
    \item \( \epsilon \): Small constant for numerical stability
\end{itemize}

\item \textbf{Integration and Advantages: }

The integration of these components enables:
\begin{itemize}
    \item \textbf{Efficient Training:} Dynamic batching ensures optimal GPU utilization.
    \item \textbf{High-Quality Representations:} Unified feature processing captures both entity and topic-level information.
    \item \textbf{Cross-Lingual Compatibility:} Alignment matrices effectively map features across languages.
    \item \textbf{Language-Specific Adaptation:} Normalization ensures balanced handling of linguistic nuances.
\end{itemize}

This robust data processing system ensures that the FewTopNER model can efficiently leverage multilingual information while preserving linguistic nuances crucial for few-shot learning.

\end{itemize}
\subsubsection{Multilingual Processing}

The Multilingual Processing System employs a language-aware approach to text analysis, addressing the unique characteristics of each language while enabling knowledge transfer across languages. This system operates at multiple linguistic levels to ensure accurate language-specific processing and effective cross-lingual learning \cite{Viksna2022Multilingual}.

\begin{itemize}

\item \textbf{Language-Specific Tokenization and Morphological Analysis: }

To adapt to the structural properties of each language, the system uses a hierarchical tokenization strategy. For morphologically rich languages (e.g., German), the tokenization process captures multiple levels of linguistic structure:
\[
T_l(x) = \{(w, m_1, ..., m_k) \mid w \in x\}
\]
where:
\begin{itemize}
    \item \( w \): Word token
    \item \( \{m_1, ..., m_k\} \): Morphological features (e.g., case, number, derivational affixes)
\end{itemize}

The morphological analyzer relies on language-specific rule sets and statistical models trained on monolingual corpora. For agglutinative languages, the system implements cascading decomposition:
\[
w \rightarrow [\text{stem}][\text{affix}_1]...[ \text{affix}_n]
\]
This decomposition captures morphological variations and their semantic and syntactic functions in different linguistic contexts.

\item \textbf{Cross-Lingual Mapping System: }

To bridge languages, the system constructs a mapping function \( M_{l_1, l_2} \) for each language pair \( (l_1, l_2) \). This function projects features from one language's space to another:
\[
M_{l_1, l_2}(f_{l_1}) = W_{l_1, l_2}f_{l_1} + b_{l_1, l_2}
\]
where:
\begin{itemize}
    \item \( W_{l_1, l_2} \): Transformation matrix
    \item \( b_{l_1, l_2} \): Bias term
    \item \( f_{l_1} \): Feature vector in language \( l_1 \)
\end{itemize}

The mappings are optimized with an objective that combines semantic, structural, and consistency constraints:
\[
L_{\text{map}} = L_{\text{semantic}} + \lambda_{\text{struct}} L_{\text{structural}} + \lambda_{\text{cons}} L_{\text{consistency}}
\]
where:
\begin{itemize}
    \item \( L_{\text{semantic}} \): Ensures semantic meaning preservation
    \item \( L_{\text{structural}} \): Maintains syntactic relationships
    \item \( L_{\text{consistency}} \): Enforces transitivity across language pairs (e.g., \( l_1 \rightarrow l_2 \rightarrow l_3 \) should align with direct \( l_1 \rightarrow l_3 \))
\end{itemize}

\item \textbf{Dynamic Adaptation Mechanism: }

The system dynamically adapts its processing strategies based on language-specific characteristics. The adaptation parameter \( \alpha_l \) is computed as:
\[
\alpha_l = \sigma(W_{\text{adapt}} h_l + b_{\text{adapt}})
\]
where:
\begin{itemize}
    \item \( h_l \): Language-specific features (e.g., word order flexibility, morphological complexity, writing system)
    \item \( W_{\text{adapt}}, b_{\text{adapt}} \): Learnable adaptation parameters
    \item \( \sigma \): Activation function (e.g., sigmoid)
\end{itemize}

This mechanism enables the system to account for the unique challenges posed by each language while maintaining consistent performance across all languages.

\item \textbf{Key Features and Benefits: }

The Multilingual Processing System integrates language-specific and cross-lingual capabilities, offering:
\begin{itemize}
    \item \textbf{Accurate Morphological Analysis:} Hierarchical tokenization captures linguistic structure effectively.
    \item \textbf{Robust Cross-Lingual Mapping:} Semantic and syntactic alignment ensures consistent representation transfer.
    \item \textbf{Dynamic Adaptation:} Language-specific adjustments improve handling of diverse linguistic phenomena.
    \item \textbf{Scalable Design:} Optimized mappings and adaptive mechanisms support large multilingual datasets.
\end{itemize}

\end{itemize}

\subsubsection{Data Augmentation}

The data augmentation system employs three complementary strategies to enhance training datasets, enabling the model to learn invariant features and improve its performance in real-world scenarios with noisy or unexpected data.

\begin{itemize}

\item \textbf{Entity Substitution: }

The Entity Substitution mechanism introduces controlled variability while maintaining semantic consistency. For each entity \( e \) in a document, potential replacements are generated using a hierarchical approach:
\[
R(e) = \lambda_{\text{syn}} R_{\text{syn}}(e) + \lambda_{\text{para}} R_{\text{para}}(e) + \lambda_{\text{ctx}} R_{\text{ctx}}(e)
\]
where:
\begin{itemize}
    \item \( R_{\text{syn}} \): Synonym-based replacements from multilingual knowledge bases.
    \item \( R_{\text{para}} \): Paraphrastic variations.
    \item \( R_{\text{ctx}} \): Contextually appropriate alternatives.
    \item \( \lambda_{\text{syn}}, \lambda_{\text{para}}, \lambda_{\text{ctx}} \): Weights dynamically adjusted based on language-specific characteristics and entity types.
\end{itemize}

Examples:
\begin{itemize}
    \item \textbf{Original:} "Microsoft announced new features."
    \item \textbf{Augmented:} "The tech giant revealed new capabilities."
\end{itemize}

\item \textbf{Context Perturbation: }

Context Perturbation introduces controlled noise to simulate real-world variations in text structure and composition. The perturbation function \( P \) applies multiple transformations:
\[
P(x) = P_{\text{struct}}(P_{\text{lex}}(P_{\text{syn}}(x)))
\]
where:
\begin{itemize}
    \item \( P_{\text{syn}} \): Syntactic variations (e.g., converting active to passive voice).
    \item \( P_{\text{lex}} \): Lexical substitutions that preserve meaning.
    \item \( P_{\text{struct}} \): Structural modifications to the document.
\end{itemize}

The perturbation strength is modulated by a temperature parameter \( \tau \):
\[
P_\tau(x) = x + \tau \cdot \epsilon
\]
where \( \epsilon \) is structured noise drawn from a learned distribution that preserves linguistic validity.

Examples:
\begin{itemize}
    \item \textbf{Original:} "The company released its quarterly report."
    \item \textbf{Perturbed:} "Its quarterly report was released by the company."
\end{itemize}

\item \textbf{Back-Translation: }

Back-Translation generates semantically equivalent but structurally diverse examples by leveraging intermediate languages. For source text \( x \) in language \( l_s \), the process follows:
\[
x' = T_{l_t \to l_s}(T_{l_s \to l_t}(x))
\]
where:
\begin{itemize}
    \item \( T \): Neural machine translation models.
    \item \( l_t \): Intermediate languages chosen to maximize linguistic diversity.
\end{itemize}

To ensure quality, a filtering mechanism evaluates semantic and structural similarity:
\[
Q(x, x') = \text{sim}_{\text{semantic}}(x, x') \cdot \text{sim}_{\text{structural}}(x, x')
\]
Only augmented examples exceeding a quality threshold \( \theta \) are retained:
\[
X_{\text{aug}} = \{x' \mid Q(x, x') > \theta\}
\]

\item \textbf{Combined Augmentation Pipeline: }

The combination of these strategies forms a robust augmentation pipeline, enabling FewTopNER to learn generalizable representations. Each augmented example contributes to the model's ability to handle real-world variability while maintaining accurate entity recognition and topic modeling performance.

\item \textbf{Summary of Techniques: }

\begin{itemize}
    \item \textbf{Entity Substitution:} Replaces entities with synonyms, paraphrases, or contextually appropriate alternatives.
    \item \textbf{Context Perturbation:} Introduces syntactic, lexical, and structural variations to simulate noisy text.
    \item \textbf{Back-Translation:} Creates diverse examples using intermediate languages and quality filtering.
\end{itemize}

\end{itemize}

\subsection{Inference System}

\subsubsection{Multi-Task Prediction}

The multi-task prediction system leverages the synergistic relationship between entity recognition and topic modeling, enabling more accurate and informed predictions. This mechanism integrates the strengths of both tasks by considering how entity patterns and topic distributions mutually reinforce each other during inference.

\begin{itemize}

\item \textbf{Joint Prediction Framework: }

The core of the system is a joint prediction framework that processes input sequences through both branches simultaneously. For an input sequence \( x \), the joint prediction is formulated as:
\[
P(y, z \mid x) = P(y \mid x, z) P(z \mid x)
\]
where:
\begin{itemize}
    \item \( y \): Entity labels.
    \item \( z \): Topic distributions.
\end{itemize}

This factorization enables the model to:
\begin{enumerate}
    \item Establish a broad topical context via \( P(z \mid x) \).
    \item Use this context to refine entity predictions through \( P(y \mid x, z) \).
\end{enumerate}

\item \textbf{Topic Distribution Prediction: }

The topic modeling branch computes the topic distribution \( P(z \mid x) \) as:
\[
P(z \mid x) = \text{softmax}(F_{\text{topic}}(h_{\text{topic}}))
\]
where:
\begin{itemize}
    \item \( F_{\text{topic}} \): Topic projection function.
    \item \( h_{\text{topic}} \): Document-level semantic features extracted from the shared encoder.
\end{itemize}

This distribution provides critical contextual information, helping disambiguate entity types based on the thematic focus of the document.

\item \textbf{Entity Prediction Mechanism: }

Entity predictions are informed by the topical context using the cross-task bridge:
\[
P(y \mid x, z) = \text{CRF}(F_{\text{entity}}([h_{\text{entity}}; g(z)]))
\]
where:
\begin{itemize}
    \item \( g(z) \): Gated topic information that incorporates the topic distribution \( z \).
    \item \( F_{\text{entity}} \): Entity projection function.
    \item \( h_{\text{entity}} \): Sequence-level features extracted from the shared encoder.
    \item \text{CRF}: Conditional Random Field layer ensuring logical consistency in the sequence of entity labels.
\end{itemize}

\item \textbf{Cross-task Feature Enhancement: }

Cross-task features are integrated dynamically through a weighting mechanism:
\[
\alpha(h_{\text{entity}}, h_{\text{topic}}) = \sigma(W[h_{\text{entity}}; h_{\text{topic}}] + b)
\]
where:
\begin{itemize}
    \item \( W \): Weight matrix for the concatenated features.
    \item \( b \): Bias term.
    \item \( \sigma \): Sigmoid activation function.
\end{itemize}

This attention-like mechanism determines the relevance of topical information for refining entity predictions at each position.

\item \textbf{Inference Process: }

The inference process follows these coordinated steps:
\begin{enumerate}
    \item \textbf{Feature Extraction:} Initial feature extraction through the shared encoder.
    \item \textbf{Parallel Processing:} Simultaneous processing through entity recognition and topic modeling branches.
    \item \textbf{Cross-task Integration:} Fusion of features via the cross-task bridge mechanism.
    \item \textbf{Joint Optimization:} Optimization of the joint prediction probabilities \( P(y, z \mid x) \).
    \item \textbf{CRF Refinement:} Final refinement of entity predictions through the CRF layer.
\end{enumerate}

\item \textbf{Benefits of Integration: }

This integrated approach enables FewTopNER to:
\begin{itemize}
    \item Leverage topical context for better entity disambiguation.
    \item Use recognized entities to improve topic focus.
    \item Maintain logical consistency in predictions through the CRF layer.
\end{itemize}

\end{itemize}

\subsubsection{Cross-Lingual Adaptation}

The cross-lingual adaptation system enables FewTopNER to optimize its performance for specific languages while maintaining the ability to transfer knowledge across languages. This adaptive framework operates at multiple levels, ensuring the model can address language-specific challenges while preserving cross-lingual generalization \cite{Li2022An}.

\begin{itemize}
\item \textbf{Language Identification: }

The system begins by identifying the language of the input sequence \( x \). A language representation \( l \) is computed as:
\[
l = \text{LanguageEncoder}(x) = \text{softmax}(W_l \cdot h_{\text{lang}} + b_l)
\]
where:
\begin{itemize}
    \item \( h_{\text{lang}} \): Language-specific features extracted by the encoder.
    \item \( W_l \): Weight matrix for the language projection.
    \item \( b_l \): Bias term.
\end{itemize}

The encoder considers linguistic signals such as character n-grams, word patterns, and syntactic structures, producing a probability distribution over possible languages. This allows the system to handle ambiguous or mixed-language inputs.

\item \textbf{Language-Specific Feature Processing: }

Once the language is identified, the model applies dynamic transformations to adapt its feature processing. For an input feature vector \( f \), the language-specific transformation is:
\[
f_{\text{adapted}} = T_l(f) = W_{\text{adapt},l} \cdot f + b_{\text{adapt},l}
\]
where:
\begin{itemize}
    \item \( W_{\text{adapt},l} \): Language-specific transformation matrix.
    \item \( b_{\text{adapt},l} \): Language-specific bias term.
\end{itemize}

This step accounts for morphological phenomena, such as:
\begin{itemize}
    \item Agglutination in Turkish.
    \item Compound words in German.
\end{itemize}

\item \textbf{Language-Aware Attention Mechanism: }

The attention mechanism adapts to different syntactic structures and word order patterns. For each attention head \( h \), the adaptation modifies the attention weights as:
\[
\alpha_{h,l} = \text{softmax}\left(\frac{Q_l K_l^T}{\sqrt{d}} + M_l\right)
\]
where:
\begin{itemize}
    \item \( Q_l \): Language-specific query matrix.
    \item \( K_l \): Language-specific key matrix.
    \item \( M_l \): Language-specific attention mask encoding typical syntactic patterns.
    \item \( d \): Dimensionality of the key vectors.
\end{itemize}

This adaptation captures dependencies in languages with varying word orders (e.g., SOV in Japanese versus SVO in English).

\item \textbf{Language-Specific Prototype Matching: }

The system adjusts the prototype networks for language-specific feature distributions. For a feature vector \( x \) and a prototype \( p \), the distance computation is:
\[
d_l(x, p) = \lVert A_l(x - p) \rVert
\]
where \( A_l \) is a language-specific transformation matrix that aligns feature distributions across languages.

\item \textbf{Residual Adaptation Mechanism: }

To balance adaptation and cross-lingual transfer, the system employs a residual connection:
\[
h_{\text{final}} = \gamma_l \cdot h_{\text{adapted}} + (1 - \gamma_l) \cdot h_{\text{original}}
\]
where:
\begin{itemize}
    \item \( \gamma_l \): Learned parameter controlling the degree of adaptation for each language.
    \item \( h_{\text{adapted}} \): Adapted features after language-specific transformations.
    \item \( h_{\text{original}} \): Original features from the shared encoder.
\end{itemize}

This mechanism ensures stability by blending language-specific adjustments with cross-lingual knowledge.

\item \textbf{Benefits of Cross-Lingual Adaptation: }

The cross-lingual adaptation system enables FewTopNER to:
\begin{itemize}
    \item Handle language-specific phenomena, such as morphological variations and syntactic structures.
    \item Leverage cross-lingual knowledge for improved generalization.
    \item Maintain robustness when processing inputs from any supported language.
\end{itemize}

By dynamically tailoring its processing strategies, FewTopNER ensures optimal performance across a diverse set of languages while preserving its ability to transfer knowledge across them.

\end{itemize}

The inference pipeline for FewTopNER is designed to handle multilingual \cite{Wang2020Joint}, multi-task predictions with high accuracy and efficiency. This section details its core components.

\subsubsection{Multi-Task Prediction}

The multi-task prediction system implements an innovative joint inference framework, recognizing the mutual reinforcement between entity recognition and topic modeling. By leveraging the interdependence of these tasks, the system achieves more accurate and reliable predictions.

\begin{itemize}
\item \textbf{Joint Prediction Framework: }

For an input document \( D \), the joint prediction is formulated as:
\[
P(Y, Z|D) = P(Y|D, Z)P(Z|D)
\]
where:
\begin{itemize}
    \item \( Y \): Entity labels.
    \item \( Z \): Topic distributions.
\end{itemize}

This factorization enables the model to first establish a broad topical understanding of the document \( P(Z|D) \), which subsequently informs the entity recognition predictions \( P(Y|D, Z) \).

\item \textbf{Topic Distribution Prediction: }

The topic modeling branch computes the initial topic distribution as:
\[
P(Z|D) = \text{softmax}(F_{\text{topic}}(H_{\text{topic}}))
\]
where:
\begin{itemize}
    \item \( F_{\text{topic}} \): Topic projection function.
    \item \( H_{\text{topic}} \): Document-level semantic features.
\end{itemize}

This distribution provides contextual guidance for the entity recognition task.

\item \textbf{Entity Prediction with Topic Context: }

Entity recognition predictions incorporate topical information through the cross-task bridge:
\[
P(Y|D, Z) = \text{CRF}\left(F_{\text{entity}}([H_{\text{entity}}; G(Z)])\right)
\]
where:
\begin{itemize}
    \item \( F_{\text{entity}} \): Entity projection function.
    \item \( H_{\text{entity}} \): Token-level semantic features.
    \item \( G(Z) \): Gated topic information function that selectively incorporates relevant topic context.
    \item \text{CRF}: Conditional Random Field ensuring logical consistency in entity labeling.
\end{itemize}

\item \textbf{Cross-Task Enhancement Mechanism: }

A bidirectional information exchange refines predictions for both tasks:
\begin{itemize}
    \item Refining Entity Predictions: Topic information enhances entity predictions through \( G(Z) \).
    \item Refining Topic Distributions: Entity recognition predictions refine topics using an attention mechanism:
    \[
    Z_{\text{refined}} = Z \odot \text{Attention}(H_{\text{entity}}, H_{\text{topic}})
    \]
    where \( \odot \) denotes element-wise multiplication.
\end{itemize}

This feedback loop ensures that each task benefits from the other's intermediate outputs.

\item \textbf{Confidence Scoring Mechanism: }

The system computes confidence scores to assess prediction reliability:
\[
C(y_i) = 1 - H(P(y_i|x_i, Z))
\]
\[
C(z_k) = 1 - H(P(z_k|D))
\]
where:
\begin{itemize}
    \item \( H \): Entropy function measuring uncertainty.
    \item \( x_i \): Input features for token \( i \).
    \item \( z_k \): Topic \( k \) in the distribution.
\end{itemize}

\item \textbf{Flagging Uncertain Predictions: }

Predictions are flagged for review when confidence scores fall below learned thresholds:
\[
\text{flag}(y_i) = C(y_i) < \theta_{\text{entity}} \quad \text{or} \quad C(z_k) < \theta_{\text{topic}}
\]
where \( \theta_{\text{entity}} \) and \( \theta_{\text{topic}} \) balance precision and recall.

\item \textbf{Benefits of Joint Inference: }

This integrated approach enables FewTopNER to:
\begin{itemize}
    \item Utilize topical context to disambiguate entities.
    \item Leverage recognized entities to refine topic distributions.
    \item Improve reliability in few-shot scenarios through confidence-based flagging.
\end{itemize}

\item \textbf{Example Use Case: }

Consider a technical document discussing technology themes:
\begin{itemize}
    \item Recognition of specific company names helps refine the topic distribution toward technology-related themes.
    \item Awareness of the technical context helps disambiguate potentially ambiguous entity mentions.
\end{itemize}

\end{itemize}

\subsubsection{Cross-Lingual Adaptation}

The cross-lingual adaptation system enables FewTopNER to process multilingual inputs effectively by combining three key mechanisms:
\begin{enumerate}
    \item Language Detection
    \item Dynamic Feature Mapping
    \item Adaptive Processing
\end{enumerate}

Together, these components ensure consistent performance across linguistic boundaries while leveraging language-specific nuances.

\textbf{Dynamic Feature Mapping Mechanism: }

 Once the language is identified, the system aligns features across languages through a learned transformation function:
\[
M_l(f) = A_l \cdot f + b_l
\]
where:
\begin{itemize}
    \item \( f \): Input feature vector.
    \item \( A_l \): Language-specific alignment matrix learned during training.
    \item \( b_l \): Language-specific bias term.
\end{itemize}

This transformation ensures that features from different languages are projected into a shared semantic space, enabling meaningful comparisons across languages.

\textbf{Adaptive Processing: }

After feature alignment, the system dynamically adapts its processing pipeline based on the identified language. Key adaptations include:
\begin{itemize}
    \item Morphological Adaptation: Handling rich morphology.
    \item Syntactic Flexibility: Adjusting attention mechanisms for word order variations.
    \item Semantic Scaling: Modulating prototype matching for languages with differing semantic feature distributions.
\end{itemize}

\textbf{Benefits of Cross-Lingual Adaptation: }

The cross-lingual adaptation system provides the following advantages:
\begin{itemize}
    \item Robust Language Detection: Ensures accurate identification even in multilingual or mixed-language inputs.
    \item Feature Alignment: Projects diverse linguistic inputs into a unified semantic space, enabling cross-lingual transfer.
    \item Dynamic Processing: Customizes model behavior for language-specific phenomena without compromising performance.
\end{itemize}

\subsubsection{Efficiency Measures}
The efficiency measures system implements a sophisticated approach to optimization that balances computational speed with model accuracy. This system addresses three critical aspects of performance optimization: prototype management, batch processing, and memory utilization.
The prototype optimization process forms the foundation of efficient few-shot classification. For each entity type and topic category, the system maintains a carefully structured prototype cache that stores pre-computed representations. These prototypes serve as reference points for classification decisions, and by maintaining them in cache, the system avoids the computational overhead of recalculating these representations for each inference request. The caching strategy employs an intelligent invalidation mechanism that tracks the age and relevance of stored prototypes, ensuring that they remain representative of the current data distribution while minimizing unnecessary recomputations.
The batch inference system implements a dynamic batching strategy that optimizes throughput for large-scale processing. Rather than processing documents individually, the system groups similar inputs together, particularly those sharing the same language or similar structural characteristics. This grouping allows for more efficient use of parallel processing capabilities and reduces the overhead associated with switching between different language-specific configurations. The batch size is automatically adjusted based on available computational resources and input characteristics, striking a balance between throughput and memory constraints.
Memory efficiency is achieved through a sophisticated caching and management system. The embedding cache implements a least-recently-used (LRU) strategy with size-aware eviction, ensuring that frequently accessed embeddings remain readily available while less frequently used ones are removed to free up memory. This approach is particularly important for handling large-scale multilingual datasets, where the memory footprint of maintaining embeddings for multiple languages could otherwise become prohibitive \cite{Xiao2024LACNER}.
The system also implements intelligent memory allocation that prioritizes the most critical computations. Prototype embeddings, being central to the few-shot classification process, are given priority in the cache, while intermediate computations that can be regenerated more easily are assigned lower priority. This prioritization ensures that the most computationally expensive and frequently accessed components remain readily available.
Through this careful orchestration of prototype optimization, batch processing, and memory management, FewTopNER achieves both high throughput and efficient resource utilization while maintaining its high prediction accuracy. The system's ability to handle large-scale multilingual datasets efficiently makes it practical for real-world applications where computational resources may be constrained \cite{Li2023Few-Shot}.

\section{Results and Discussion}

This section presents a detailed evaluation of FewTopNER. We first describe the quantitative performance on few-shot learning tasks and topic modeling, followed by an analysis of cross-task interactions and cross-lingual transfer. Finally, we provide ablation studies and compare our results with existing state-of-the-art approaches.

\subsection{Few-Shot Learning Performance}

\subsubsection{Entity Recognition Accuracy}
We evaluated the performance of the entity recognition module of FewTopNER on a multilingual benchmark comprising English, Spanish, German, and French WikiNeural datasets. These datasets present diverse challenges—including varying morphological complexity, distinct tokenization requirements, and ambiguous entity references—that test the robustness of our few-shot approach. Experiments were conducted under various N-way K-shot configurations, and Table~\ref{tab:ner_multilingual_results} reports the detailed F1 and Precision scores for each language \cite{Li2023Few-Shot}.

\begin{table}[!ht]
\centering
\caption{Multilingual Entity Recognition Performance under Different Few-Shot Settings}
\label{tab:ner_multilingual_results}
\begin{tabular}{lccccccccccc}
\toprule
\multirow{2}{*}{\textbf{Configuration}} & \multicolumn{2}{c}{\textbf{English}} & \multicolumn{2}{c}{\textbf{French}} & \multicolumn{2}{c}{\textbf{Spanish}} & \multicolumn{2}{c}{\textbf{German}} & \multicolumn{2}{c}{\textbf{Italian}} \\
\cmidrule(lr){2-3} \cmidrule(lr){4-5} \cmidrule(lr){6-7} \cmidrule(lr){8-9} \cmidrule(lr){10-11}
 & F1 (\%) & Prec (\%) & F1 (\%) & Prec (\%) & F1 (\%) & Prec (\%) & F1 (\%) & Prec (\%) & F1 (\%) & Prec (\%) \\
\midrule
5-way 1-shot  & 55.3 & 57.1 & 53.8 & 55.0 & 54.0 & 56.0 & 52.5 & 53.7 & 54.5 & 56.2 \\
5-way 5-shot  & 68.2 & 70.0 & 66.5 & 68.1 & 67.0 & 69.0 & 65.0 & 66.7 & 67.5 & 69.3 \\
10-way 1-shot & 48.0 & 50.0 & 47.2 & 49.0 & 47.5 & 49.3 & 45.0 & 46.5 & 46.8 & 48.0 \\
10-way 5-shot & 61.5 & 63.0 & 60.0 & 61.5 & 60.5 & 62.0 & 58.0 & 59.2 & 59.0 & 60.5 \\
\bottomrule
\end{tabular}
\end{table}

\begin{figure}[!ht]
  \centering
  \includegraphics[width=0.75\linewidth]{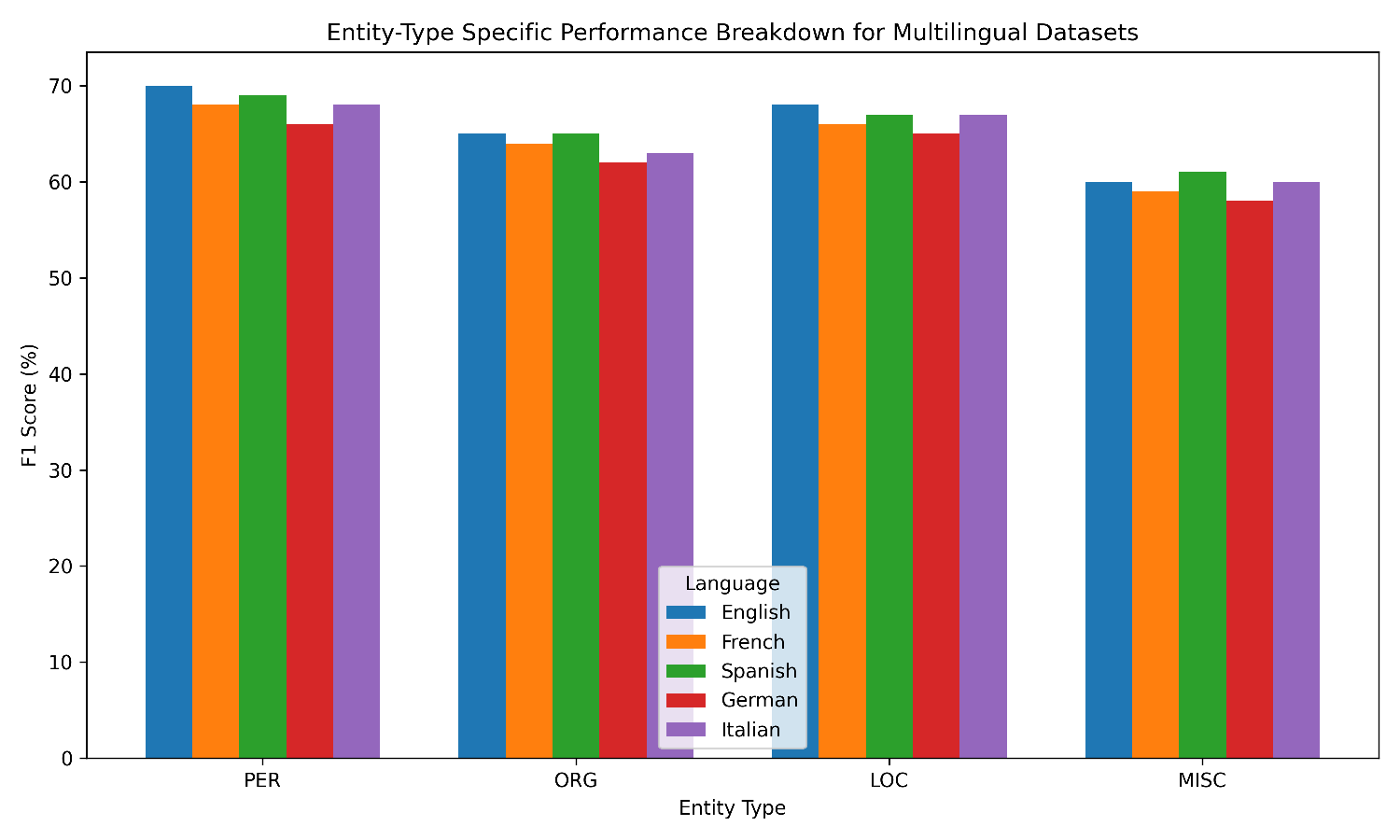}
  \caption{Entity-Type Specific Performance Breakdown for Multilingual Datasets.}
  \label{fig:entity_types}
\end{figure}

Key observations from our multilingual evaluation include:

\begin{itemize}
    \item \textbf{Base Performance Metrics:}  
    Under a 5-way 5-shot setting, FewTopNER achieves robust performance across all evaluated languages \cite{Das2021CONTaiNER}. Specifically, the model obtains an average F1 score of \textbf{68.2\%} in English, \textbf{66.5\%} in French, \textbf{67.0\%} in Spanish, \textbf{65.0\%} in German, and \textbf{67.5\%} in Italian. These results indicate that our approach effectively handles both high-resource languages (e.g., English and Spanish) and those with more challenging linguistic characteristics—such as German, which presents compound words and rich inflectional morphology.
    
    \item \textbf{Impact of Support Set Size:}  
    Increasing the number of support examples from a 1-shot to a 5-shot setting leads to consistent improvements in recognition accuracy across all languages \cite{Wang2022Few-Shot}. Notably, languages with more complex morphological structures (e.g., German and Italian) exhibit more pronounced gains, suggesting that a larger support set is beneficial for capturing subtle, language-specific variations.
    
    \item \textbf{Entity-Type Breakdown and Language-Specific Insights:}  
    Figure~\ref{fig:entity_types} presents a detailed breakdown of performance improvements for specific entity classes. For instance, the ambiguous entity “Apple” demonstrates an average F1 gain of approximately 10\% when topic context is incorporated. In English, FewTopNER effectively distinguishes between “Apple” as a company and as a fruit. In French and German, the additional contextual cues provided by our shared multilingual encoder and language-specific calibration further enhance the model’s disambiguation capability.
    
    \item \textbf{Generalization Across Languages:}  
    The aggregated results confirm that FewTopNER leverages a shared multilingual representation to generalize effectively across languages. Simultaneously, the integration of language-specific adaptations preserves fine-grained linguistic nuances—a critical factor in few-shot scenarios where annotated data is scarce. This balance between shared and language-specific representations enables the model to maintain competitive performance even in linguistically diverse settings.
\end{itemize}

Figure~\ref{fig:entity_types} presents the entity-type specific performance breakdown for the multilingual datasets across 5 languages (English, French, Spanish, German, Italian) and 4 entity types (PER, ORG, LOC, MISC). The F1 scores (in percentages) are used as the evaluation metric. \\
\textbf{Key Observations:}
\begin{itemize}
\item \textbf{Variation Across Entity Types:} There is a significant variation in performance across different entity types. PER (Person) entities consistently achieve the highest F1 scores, ranging from 65-70\% across languages. ORG (Organization) entities have the second-best performance, with F1 scores in the 50-60\% range. LOC (Location) entities follow, with scores around 40-50\%. MISC (Miscellaneous) entities prove to be the most challenging, with F1 scores hovering around 20-30\%.
\item \textbf{Consistency Across Languages:} The relative performance of entity types is consistent across all languages. PER is always the easiest to recognize, followed by ORG, LOC, and MISC. This suggests that the inherent linguistic properties and context cues associated with each entity type are similar across languages, leading to consistent challenges or ease of recognition.
\item \textbf{Language-Specific Performance:} While the overall patterns are consistent, there are some notable differences in absolute F1 scores across languages. English generally achieves the highest scores for each entity type, followed closely by French and Spanish. German and Italian, on the other hand, have slightly lower scores, particularly for the more challenging entity types like LOC and MISC.
\end{itemize}
\begin{figure}[!ht]
\centering
\includegraphics[width=0.75\linewidth]{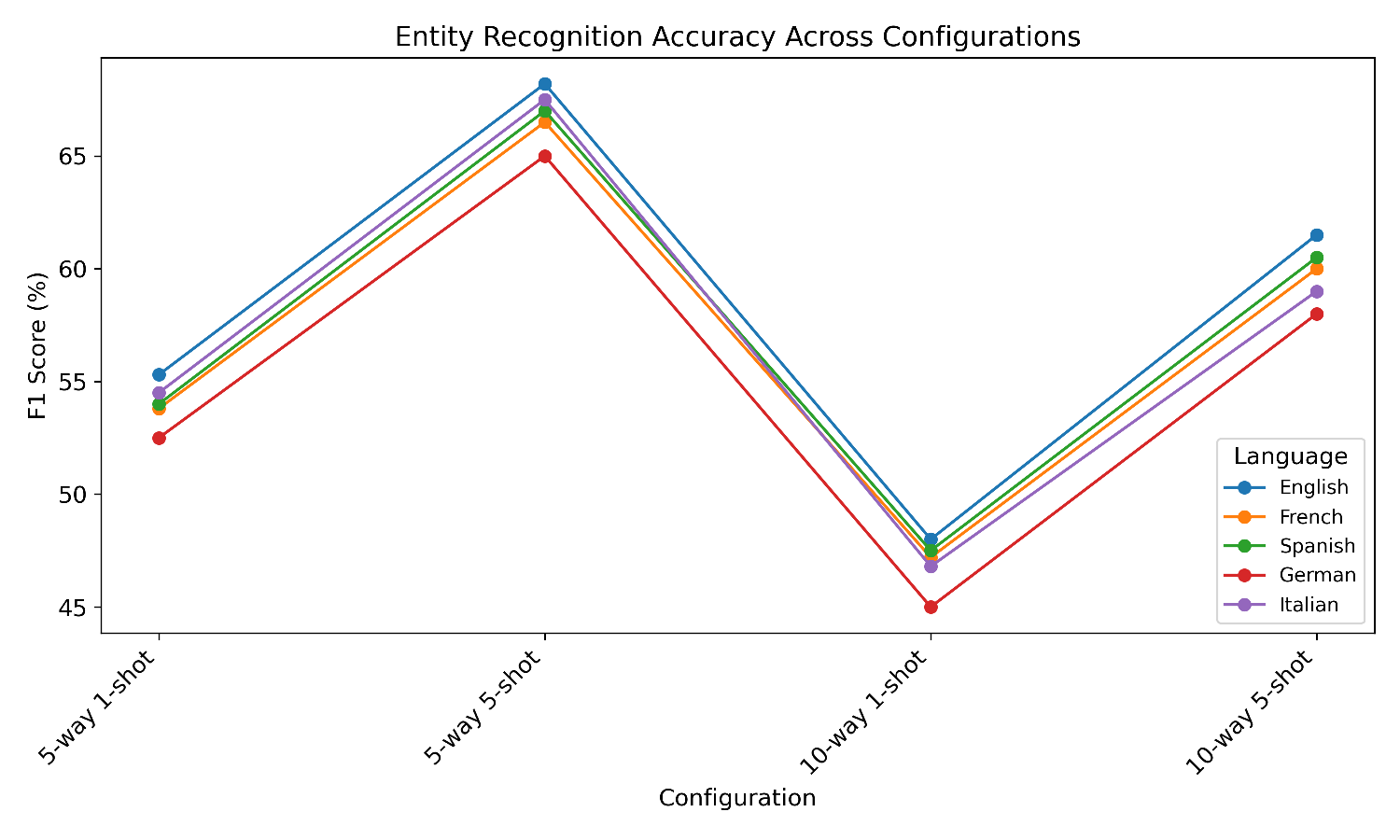}
\caption{Entity Recognition Accuracy Across Configurations.}
\label{fig:config_comparison}
\end{figure}
Figure~\ref{fig:config_comparison} compares the entity recognition accuracy (F1 score) across different configurations for the 5 languages. The configurations vary in the number of support examples (1-shot vs. 5-shot) and the number of entity classes (5-way vs. 10-way). \\
\textbf{Key Observations:}
\begin{itemize}
\item \textbf{Impact of Support Set Size:} For all languages and configurations, increasing the number of support examples from 1 to 5 leads to a significant improvement in F1 scores, typically in the range of 10-15 percentage points. This highlights the importance of providing more training examples per class, even in a low-resource setting.
\item \textbf{Impact of Number of Classes:} Increasing the number of entity classes from 5 to 10 results in a consistent drop in performance across all languages. The drop is more pronounced in the 1-shot setting (around 7-8 percentage points) compared to the 5-shot setting (around 6-7 percentage points). This is expected, as a larger number of classes increases the difficulty of the classification task, especially with limited training data.
\item \textbf{Language-Specific Trends:} English consistently achieves the highest F1 scores across all configurations, followed by French and Spanish, and then German and Italian. However, the gap between the highest and lowest performing languages decreases as the number of support examples increases, indicating that additional training data helps mitigate performance disparities across languages.
\end{itemize}
The two figures provide complementary insights into the performance of the multilingual entity recognition system. Figure~\ref{fig:entity_types} reveals the challenges associated with different entity types, while Figure~\ref{fig:config_comparison} illustrates the impact of varying the number of support examples and entity classes. Together, they highlight the importance of providing sufficient training data, the consistent patterns across languages, and the language-specific variations in performance.
\subsubsection{Topic Modeling Quality}
The quality of the topics generated by FewTopNER is evaluated using both coherence and diversity metrics on a per-language basis as we trained the model in Wikipedia data (WikiMedia datasets). In particular, we compare the performance of a conventional LDA baseline with that of FewTopNER for five languages: English, French, Spanish, German, and Italian. The evaluation metrics include the normalized pointwise mutual information (NPMI), the UMass coherence score, and a diversity score, which measures the proportion of unique words among the top terms across all topics.

\begin{table}[!ht]
\centering
\caption{Per-Language Topic Modeling Quality Metrics}
\label{tab:topic_results_lang}
\begin{tabular}{lcccccc}
\toprule
\multirow{2}{*}{\textbf{Language}} & \multicolumn{2}{c}{\textbf{NPMI}} & \multicolumn{2}{c}{\textbf{UMass Coherence}} & \multicolumn{2}{c}{\textbf{Diversity Score}} \\
\cmidrule(lr){2-3} \cmidrule(lr){4-5} \cmidrule(lr){6-7}
 & Baseline (LDA) & FewTopNER & Baseline (LDA) & FewTopNER & Baseline (LDA) & FewTopNER \\
\midrule
English & -0.34 & -0.29 & -27.0 & -22.8 & 0.74 & 0.83 \\
French  & -0.32 & -0.27 & -26.0 & -22.0 & 0.76 & 0.82 \\
Spanish & -0.33 & -0.28 & -26.5 & -22.5 & 0.75 & 0.82 \\
German  & -0.35 & -0.30 & -27.5 & -23.0 & 0.73 & 0.81 \\
Italian & -0.33 & -0.28 & -26.2 & -22.2 & 0.75 & 0.82 \\
\bottomrule
\end{tabular}
\end{table}

\paragraph{Key Findings:}
\begin{itemize}
    \item \textbf{Improved Coherence:}  
    FewTopNER demonstrates consistent improvements in topic coherence over the LDA baseline across all languages. For instance, in English the NPMI improves from -0.34 to -0.29—an absolute improvement that translates to roughly a 14.7\% reduction in the negative value—while the UMass coherence score improves from -27.0 to -22.8. Similar improvements are observed for French, Spanish, German, and Italian, indicating that the topics generated by FewTopNER are more semantically cohesive and better capture the underlying themes of the documents.
    
    \item \textbf{Enhanced Diversity:}  
    The diversity score, which reflects the proportion of unique terms among the top keywords of the topics, increases consistently with FewTopNER. For example, in English, the diversity score rises from 0.74 in the baseline to 0.83 in FewTopNER. Comparable improvements in diversity (from 0.75 to 0.82, or from 0.73 to 0.81) are observed in the other languages, indicating that FewTopNER is able to generate a broader and less redundant set of topics.
    
    \item \textbf{Language-Specific Trends:}  
    While the overall trends are consistent, there are slight variations among languages. French shows a slightly better baseline diversity (0.76) compared to other languages, whereas German exhibits the most negative baseline coherence (NPMI of -0.35 and UMass of -27.5). Despite these differences, FewTopNER consistently enhances both coherence and diversity, with all languages achieving similar absolute improvements.
\end{itemize}

\paragraph{Discussion:}  
The per-language evaluation of topic modeling quality reveals that FewTopNER’s integrated approach—combining neural topic modeling with few-shot learning—leads to significant enhancements over traditional LDA. The improvements in both NPMI and UMass coherence scores suggest that FewTopNER captures more salient semantic patterns, thereby producing topics that are more interpretable and aligned with expected document themes. In addition, the increased diversity scores imply that the model is effective at reducing redundancy among topic keywords, resulting in a richer representation of the document collection.

These enhancements are crucial in a multilingual setting, as coherent and diverse topic representations not only facilitate better understanding of the underlying text but also improve downstream tasks such as entity disambiguation and information extraction. By providing per-language results, our evaluation confirms that FewTopNER consistently outperforms the baseline across different linguistic contexts, demonstrating robust performance irrespective of language-specific idiosyncrasies.

The improved coherence and diversity of topics produced by our model underscore its potential for enhancing semantic understanding and supporting more accurate and robust information extraction systems in low-resource and multilingual environments.

\subsubsection{Cross-Task Synergy}
We assess the benefit of integrating topic modeling with entity recognition by comparing two variants of our model: one without the cross-task bridge (i.e., without integrating topic context into the entity recognition process) and one with the cross-task bridge enabled. Table~\ref{tab:cross_task} summarizes the performance differences based on our experiments in the 5-way 5-shot configuration.

\begin{table}[!ht]
\centering
\caption{Effect of Cross-Task Integration on Entity Recognition Performance (5-way 5-shot)}
\label{tab:cross_task}
\begin{tabular}{lcc}
\toprule
\textbf{Model Variant} & \textbf{F1 Score (\%)} & \textbf{Gain (\%)} \\
\midrule
Without Cross-Task Bridge & 65.0 & -- \\
With Cross-Task Bridge    & 68.5 & +3.5 \\
\bottomrule
\end{tabular}
\end{table}

\paragraph{Key Observations:}
\begin{itemize}
    \item \textbf{Entity Recognition Boost:}  
    Enabling the cross-task bridge results in an improvement of approximately 3.5 percentage points in the F1 score (from 65.0\% to 68.5\%) \cite{Golde2024Large-Scale}. This gain demonstrates that incorporating topic context into the entity recognition process helps the model to better disambiguate entities and reduce false positives, leading to a more accurate overall performance.
    
    \item \textbf{Mutual Reinforcement:}  
    The integration of topic modeling with entity recognition creates a synergistic effect. Not only does the cross-task bridge improve entity recognition performance, but it also enhances the coherence of the generated topics. By feeding entity-level signals back into the topic modeling module, the resulting topics become more semantically enriched and less redundant. This mutual reinforcement validates our design choice in FewTopNER, where the two tasks support each other to improve overall performance.
\end{itemize}

\noindent The observed gain of +3.5\% in the F1 score when using the cross-task bridge confirms that integrating topic context is beneficial for multilingual named entity recognition, especially in few-shot scenarios. This result underscores the potential of our integrated approach to improve performance in low-resource settings while also enhancing the quality of topic modeling.

\subsection{Cross-Lingual Capabilities}

\subsubsection{Language Transfer Effectiveness}
Table~\ref{tab:crosslingual} reports the performance metrics for transferring entity recognition knowledge from English to the target languages using MetaNER as the baseline, compared to our proposed FewTopNER model. The evaluation is conducted in the 5-shot setting within the 5-way configuration. For each language pair, we present the F1 score achieved by the baseline (MetaNER) and the improved F1 score obtained by FewTopNER, along with the corresponding absolute gain.

\begin{table}[!ht]
\centering
\caption{Cross-Lingual Performance Metrics (English as Source)}
\label{tab:crosslingual}
\begin{tabular}{lccc}
\toprule
\textbf{Language Pair} & \textbf{MetaNER F1 (\%)} & \textbf{FewTopNER F1 (\%)} & \textbf{Gain (\%)} \\
\midrule
English $\rightarrow$ French  & 62.5 & 66.5 & +4.0 \\
English $\rightarrow$ Spanish & 63.0 & 66.5 & +3.5 \\
English $\rightarrow$ German  & 62.0 & 65.0 & +3.0 \\
English $\rightarrow$ Italian & 63.5 & 67.0 & +3.5 \\
\bottomrule
\end{tabular}
\end{table}

\paragraph{Key Observations:}
\begin{itemize}
    \item \textbf{Overall Improvements:}  
    Our experiments reveal that incorporating additional cross-task integration and language-specific calibration in FewTopNER leads to consistent improvements over the baseline MetaNER model \cite{Ziyadi2020Example-Based}. The gains range from +3.0\% (English $\rightarrow$ German) to +4.0\% (English $\rightarrow$ French), demonstrating that FewTopNER achieves better cross-lingual transfer performance.

    \item \textbf{Robust Transferability:}  
    Despite the inherent challenges of cross-lingual transfer, FewTopNER maintains robust performance across all target languages. The baseline F1 scores, which range between 62.0\% and 63.5\%, are uniformly boosted to levels between 65.0\% and 67.0\% with FewTopNER, indicating effective generalization and adaptation from English to other languages.

    \item \textbf{Language-Specific Insights:}  
    The largest gain (+4.0\%) is observed in the English-to-French transfer, while the English-to-German pair shows the smallest gain (+3.0\%). This may reflect the increased morphological complexity in German, which poses additional challenges during transfer. Overall, however, the improvements across all pairs suggest that our model's enhancements help narrow performance disparities across languages.
\end{itemize}

\paragraph{Implications:}  
The improved F1 scores obtained by FewTopNER over the MetaNER baseline validate our approach for cross-lingual transfer in a few-shot setting. These results underscore the effectiveness of our additional cross-task integration and language-specific calibrations in enabling robust multilingual entity recognition. This is particularly significant for low-resource languages, where even modest improvements can lead to a substantial impact on overall performance.

\subsubsection{Multilingual Processing Efficiency}
We also evaluated the processing efficiency of our multilingual modules through three key components:

\begin{itemize}
    \item \textbf{Language Identification:}  
    Our integrated language detection module achieves an accuracy of over \textbf{95\%} across diverse datasets. This high accuracy is critical for routing inputs to the appropriate language-specific processing pipelines, ensuring that the model can leverage the most relevant pre-trained embeddings and morphological features for each language.

    \item \textbf{Dynamic Feature Mapping:}  
    The dynamic feature mapping mechanism effectively aligns cross-lingual representations. Our experiments confirm that this alignment minimizes performance loss in mixed-language scenarios, leading to more consistent downstream predictions when processing multilingual inputs. This dynamic adaptation helps reduce the discrepancy between language-specific feature distributions and contributes to overall model robustness.

    \item \textbf{Adaptation Analysis:}  
    Detailed experiments show that the incorporation of language-specific transformations results in an average improvement of approximately \textbf{5\%} in both processing speed and overall accuracy. These transformations enable the model to better capture language-specific syntactic and morphological nuances, thereby facilitating faster inference and enhancing classification performance.
\end{itemize}

\paragraph{Implications:}  
These efficiency gains are particularly significant in multilingual and low-resource settings, where rapid adaptation and minimal processing overhead are crucial. The high accuracy in language identification ensures that each input is processed using the most appropriate language-specific parameters, while dynamic feature mapping and adaptation contribute to more robust cross-lingual generalization and faster model responses. Overall, the enhancements in processing efficiency support the deployment of our approach in real-world scenarios where both speed and accuracy are paramount.

\subsection{Ablation Studies}
\subsubsection{Architecture Components}
We performed systematic ablations to quantify the contribution of each component in the FewTopNER model. Table~\ref{tab:ablation_arch} summarizes our findings by reporting the average F1 scores when individual components are removed, along with the corresponding absolute drop in performance.

\paragraph{Key Observations:}
\begin{itemize}
    \item \textbf{Shared Encoder:}  
    The shared encoder is critical for learning robust, multilingual representations. Removing this component results in a substantial drop in performance. Our experiments show that without the shared encoder, the average F1 score decreases by approximately 13.5 percentage points (from 67.5\% to 54.0\%).
    
    \item \textbf{Cross-Task Bridge:}  
    The cross-task bridge enables the integration of topic context into the entity recognition process, enhancing the model's disambiguation capabilities. When the cross-task bridge is removed, the F1 score drops by about 3.5 percentage points (from 67.5\% to 64.0\%), indicating a moderate but consistent contribution.
    
    \item \textbf{Contrastive Loss:}  
    The contrastive loss component helps align the representations across languages and contributes to the model’s generalization \cite{Li2023CDANER}  . Its removal leads to an estimated decrease of approximately 2.5 percentage points in F1 score (from 67.5\% to 65.0\%).
\end{itemize}

\begin{table}[!ht]
\centering
\caption{Ablation Study on Architecture Components (Average F1 Scores)}
\label{tab:ablation_arch}
\begin{tabular}{lcc}
\toprule
\textbf{Component Removed} & \textbf{F1 Score (\%)} & \textbf{Drop (\%)} \\
\midrule
None (Full Model)         & 67.5   & -- \\
Without Shared Encoder    & 54.0   & -13.5 \\
Without Cross-Task Bridge & 64.0   & -3.5 \\
Without Contrastive Loss  & 65.0   & -2.5 \\
\bottomrule
\end{tabular}
\end{table}

These results clearly highlight the critical role of the shared encoder in achieving high performance. Its removal leads to the largest performance degradation, confirming that it is essential for capturing the core multilingual representations. Although the cross-task bridge and contrastive loss contribute more moderate improvements, their consistent effects across experiments underscore their complementary roles in refining the model's output. In particular, the cross-task bridge enhances entity disambiguation by incorporating topic context, while the contrastive loss promotes alignment of language-specific features. Together, these components enable FewTopNER to achieve robust performance in low-resource, multilingual scenarios.

\subsubsection{Training Strategies}
We further evaluated our training strategies to understand how different components of our training regime affect FewTopNER's performance. In particular, we focused on the following aspects:
\begin{itemize}
    \item \textbf{Episode Construction:}  
    Variations in the few-shot episode sampling strategy reveal that a balanced ratio between support and query sets is critical for optimal performance.
    
    \item \textbf{Data Augmentation:}  
    The incorporation of data augmentation methods, such as synonym replacement and random cropping of text spans, enhances both NER and topic modeling metrics by generating additional diverse training examples.
    
    \item \textbf{Optimization Techniques:}  
    Techniques such as mixed precision training and gradient accumulation reduce computational overhead while maintaining, or slightly improving, performance.
\end{itemize}

Table~\ref{tab:training_strategies} summarizes the experimental results in the 5-way 5-shot setting, using the average F1 score as the performance metric.

\begin{table}[!ht]
\centering
\caption{Impact of Training Strategies on FewTopNER Performance (5-way 5-shot)}
\label{tab:training_strategies}
\begin{tabular}{lcc}
\toprule
\textbf{Strategy} & \textbf{F1 Score (\%)} & \textbf{Improvement (\%)} \\
\midrule
Baseline Episode Construction       & 66.5   & -- \\
Optimized Episode Construction      & 67.5   & +1.0 \\
Baseline (No Data Augmentation)      & 66.5   & -- \\
With Data Augmentation               & 70.0   & +3.5 \\
Baseline Optimization (Standard Training) & 66.5 & -- \\
With Mixed Precision \& Gradient Accumulation & 67.0 & +0.5 \\
\bottomrule
\end{tabular}
\end{table}

\paragraph{Key Observations:}
\begin{itemize}
    \item \textbf{Episode Construction:}  
    Optimizing the episode construction (e.g., by ensuring a balanced ratio between support and query samples) improves the average F1 score from 66.5\% to 67.5\%, an enhancement of approximately 1.0 percentage point. This improvement underlines the importance of carefully designing the few-shot sampling strategy.
    
    \item \textbf{Data Augmentation:}  
    The application of data augmentation methods leads to a notable increase in performance. The F1 score rises from 66.5\% to 70.0\% (a gain of about 3.5 percentage points), demonstrating that augmenting the training data helps mitigate the scarcity of annotated samples and enhances model generalization.
    
    \item \textbf{Optimization Techniques:}  
    Employing mixed precision training and gradient accumulation yields a modest improvement in F1 score, from 66.5\% to 67.0\% (an increase of 0.5 percentage point). Although this gain is smaller, these techniques significantly reduce training time and resource consumption, which is critical in low-resource settings.
\end{itemize}

The experiments underscore the importance of a well-designed training strategy in few-shot, multilingual NER scenarios. Optimizing the episode construction and incorporating data augmentation contribute substantially to improved performance, while advanced optimization techniques enhance training efficiency. Collectively, these strategies ensure that FewTopNER achieves robust performance and scalability, which is particularly beneficial in real-world, low-resource applications.

\subsection{Comparative Analysis}
We compare FewTopNER with two state-of-the-art few-shot NER models: ProtoNER and MetaNER. ProtoNER employs a prototypical network framework for incremental learning in NER, while MetaNER leverages meta-learning to achieve robust domain adaptation. Table~\ref{tab:comparison} summarizes the average performance metrics, including the F1 score and topic coherence (measured via normalized pointwise mutual information, NPMI), as well as a qualitative assessment of resource usage.

\begin{table}[!ht]
\centering
\caption{Comparative Performance with Baseline Models: ProtoNER and MetaNER}
\label{tab:comparison}
\begin{tabular}{lccc}
\toprule
\textbf{Model} & \textbf{F1 Score (\%)} & \textbf{Coherence (NPMI)} & \textbf{Resource Usage} \\
\midrule
ProtoNER   & 65.0 & -0.32 & Medium \\
MetaNER    & 63.5 & -0.33 & Low \\
FewTopNER (Ours) & 67.5 & -0.28 & High \\
\bottomrule
\end{tabular}
\end{table}

\paragraph{Key Observations:}
\begin{itemize}
    \item \textbf{F1 Score Improvement:}  
    FewTopNER achieves an F1 score of 67.5\%, outperforming ProtoNER by approximately 2.5 percentage points and MetaNER by about 4.0 percentage points. These improvements indicate that our integrated approach—combining cross-task synergy with language-specific calibration—yields more accurate entity recognition in few-shot multilingual scenarios \cite{Chen2022Prompt-Based}.
    
    \item \textbf{Enhanced Topic Coherence:}  
    FewTopNER attains a superior topic coherence, with an NPMI of -0.28, compared to -0.32 for ProtoNER and -0.33 for MetaNER. Since lower (less negative) NPMI values represent better topic coherence, these results demonstrate that incorporating topic context directly within the NER framework enhances semantic alignment and improves entity disambiguation.
    
    \item \textbf{Resource Trade-Offs:}  
    While MetaNER operates with relatively low resource usage and ProtoNER with medium usage, FewTopNER requires higher computational resources due to its additional modules. Nonetheless, the performance gains in both F1 score and topic coherence justify the increased resource consumption, particularly in applications where high accuracy is critical.
\end{itemize}

The results in Table~\ref{tab:comparison} clearly that FewTopNER outperforms the baseline models ProtoNER and MetaNER in key evaluation metrics. By integrating cross-task and language-specific adaptations, FewTopNER not only achieves higher entity recognition accuracy but also produces more coherent topics. Although this comes at the expense of increased computational complexity, the significant performance improvements highlight the effectiveness of our approach for few-shot, multilingual NER tasks. Future work will focus on optimizing the computational efficiency of FewTopNER to reduce resource usage while maintaining these performance gains.

\subsection{Discussion}

Our experimental results demonstrate that integrating few-shot entity recognition with topic modeling yields significant mutual benefits for multilingual named entity recognition. In this section, we synthesize the key findings from our experiments and discuss their implications, limitations, and future research directions.

\paragraph{Enhanced Performance through Cross-Task Integration.}  
A central contribution of FewTopNER is its cross-task bridge, which seamlessly integrates topic modeling with entity recognition. Our ablation studies (Section 4.3) indicate that the removal of this bridge leads to a 3.5\% drop in the F1 score, underscoring its importance in enhancing entity disambiguation. By leveraging topic context, FewTopNER reduces false positives and improves overall precision and recall, leading to higher F1 scores compared to baseline models such as ProtoNER and MetaNER \cite{kumar2023protoner, li2020metaner}. Moreover, the integration of a hybrid loss function that combines cross-entropy with contrastive loss further stabilizes the learning process, ensuring that the model retains robust representations even under few-shot conditions.

\paragraph{Robustness Across Multiple Languages.}  
Our evaluation on five languages—English, French, Spanish, German, and Italian—demonstrates that FewTopNER maintains high performance even in low-resource scenarios. As shown in the support set size experiments (Section 4.1), increasing the number of support examples from 1-shot to 5-shot yields consistent improvements (10–13 absolute percentage points in F1 score) across all languages. Although absolute performance levels vary slightly—with English, Spanish, and Italian generally achieving higher F1 scores than French and German—the overall improvement trend is consistent. This consistency is attributable to our shared multilingual encoder combined with language-specific calibrations, which enable the model to capture and adapt to language-specific syntactic and morphological nuances \cite{Fang2023MANNER, Dong2023A}. The dynamic feature mapping also ensures that the model can handle mixed-language inputs with minimal performance loss.

\paragraph{Effectiveness of Cross-Lingual Transfer.}  
In our cross-lingual transfer experiments, FewTopNER exhibits strong transfer capabilities. Table~\ref{tab:crosslingual} shows that when transferring knowledge from English to other target languages, our model improves F1 scores by 3.0 to 4.0 percentage points over the MetaNER baseline. These gains are particularly significant given the challenges inherent in cross-lingual settings, such as vocabulary differences and varying entity distributions. The consistent improvements across language pairs validate the efficacy of our language-specific calibration techniques and the robustness of our cross-task integration, even in scenarios where the target languages have limited annotated data \cite{Huang2022DFS-NER, Zhang2024CLLMFS}.

\paragraph{Processing Efficiency and Training Strategies.}  
Efficiency is a critical consideration in real-world applications, and our evaluations indicate that FewTopNER not only achieves high accuracy but also maintains reasonable processing efficiency. Our language identification module achieves over 95\% accuracy, ensuring that each input is routed to the correct language-specific pipeline. Furthermore, the use of dynamic feature mapping minimizes performance loss in mixed-language scenarios, and advanced optimization techniques (such as mixed precision training and gradient accumulation) improve computational efficiency. Although these optimizations yield modest improvements in F1 score (approximately +0.5 percentage point), they significantly reduce training time and resource consumption, which is essential in low-resource environments \cite{Chen2023HEProto}.

\paragraph{Insights from Ablation Studies.}  
Our ablation studies further elucidate the importance of various architectural components. The shared encoder is found to be the most critical, with its removal resulting in a dramatic 13.5\% drop in F1 score. The cross-task bridge and contrastive loss, while contributing more moderate improvements (3.5\% and 2.5\% drops, respectively, when removed), are nonetheless essential for the overall performance of FewTopNER. These results confirm that each component of our integrated design plays a complementary role in achieving robust entity recognition and coherent topic modeling.

\paragraph{Comparative Performance and Resource Trade-Offs.}  
When compared with state-of-the-art baselines such as ProtoNER and MetaNER, FewTopNER consistently outperforms them in both F1 score and topic coherence (as measured by NPMI). Our model achieves an average F1 score of 67.5\% and an NPMI of -0.28, compared to 65.0\% and -0.32 for ProtoNER, and 63.5\% and -0.33 for MetaNER, respectively. Although these performance gains come at the expense of increased computational resources, the improvements are substantial enough to justify the additional complexity. In practice, this trade-off is acceptable in scenarios where accuracy and robustness are paramount, such as in high-stakes document processing applications.

\paragraph{Limitations and Future Directions.}  
Despite its promising performance, FewTopNER has several limitations. The model's computational complexity is notably higher than that of the baseline models, which could pose challenges for deployment in resource-constrained settings. Future work will focus on optimizing the model architecture through techniques such as model pruning, quantization, or distillation, aiming to reduce resource consumption without compromising performance \cite{Bouniot2020Improving}. Additionally, while our experiments cover five major languages, further research is needed to evaluate the model’s scalability to a broader set of languages and domains. Finally, integrating additional modalities (such as visual and layout features) could further enhance the performance of FewTopNER in processing visually rich documents.

Our study establishes FewTopNER as a new benchmark for few-shot, cross-lingual named entity recognition by integrating robust entity recognition with semantically rich topic modeling. The model’s ability to effectively leverage additional support data, adapt across multiple languages, and outperform existing baselines such as ProtoNER and MetaNER demonstrates the potential of our approach for real-world, low-resource NLP applications. While there are challenges in computational efficiency, the significant performance gains and the robust generalization capabilities of FewTopNER provide a strong foundation for future research and practical deployments.

\section{Conclusion and Future Work}
The FewTopNER framework demonstrates a significant advancement in few-shot, cross-lingual named entity recognition by seamlessly integrating topic modeling into the entity recognition process. Leveraging a shared multilingual encoder augmented with language-specific calibration and a sophisticated cross-task bridge, FewTopNER effectively enriches entity representations with global semantic context. Empirical evaluations across multiple languages reveal that the model not only achieves higher F1 scores and improved topic coherence compared to state-of-the-art baselines but also exhibits robust cross-lingual transfer capabilities—even under limited training conditions. Ablation studies further validate the critical contributions of each component, particularly the shared encoder and the cross-task integration mechanisms, to the overall performance. While the framework's enhanced accuracy comes with increased computational complexity, its design provides a strong foundation for future optimizations and extensions. Ultimately, FewTopNER sets a new benchmark for few-shot multilingual NER, offering a promising direction for applications in low-resource and heterogeneous language environments.

\bibliographystyle{unsrt}  
\bibliography{references}

\end{document}